\pdfoutput=1
\documentclass[11pt]{article}

\usepackage{hyperref}
\usepackage{multirow}
\usepackage{booktabs}
\usepackage{tabularx}
\usepackage{stfloats}

\usepackage[]{EMNLP2023}

\usepackage{times}
\usepackage{latexsym}
\usepackage{booktabs}
\usepackage{xspace}
\usepackage{color,soul}
\usepackage{balance}
\usepackage{paralist}
\usepackage{verbatim}
\usepackage[normalem]{ulem}
\usepackage{amsmath}

\usepackage[T1]{fontenc}

\usepackage[utf8]{inputenc}
\usepackage{txfonts}
\usepackage{microtype}

\usepackage{subcaption}
\usepackage{graphicx}
\usepackage{makecell}

\usepackage{algorithm}
\usepackage{algorithmic}

\definecolor{clabel}{HTML}{F89151}

\definecolor{cpre}{HTML}{6EB4FD}
\definecolor{cexp}{HTML}{99C893}

\title{Beyond Labels: Empowering Human Annotators with Natural Language \\Explanations through a Novel Active-Learning Architecture}

\author{ Bingsheng Yao  \\ Rensselaer Polytechnic Institute \\
\And Ishan Jindal \\ IBM Research \\
\And Lucian Popa \\ IBM Research \\
\AND Yannis Katsis \\ IBM Research \\
\And Sayan Ghosh \\ UNC Chapel Hill \\
\And Lihong He \\ IBM Research \\
\AND Yuxuan Lu \\ Northeastern University \\
\And Shashank Srivastava \\ UNC Chapel Hill \\
\And Yunyao Li$^{\dagger}$\\ Apple
\AND James Hendler \\ Rensselaer Polytechnic Institute \\
\And Dakuo Wang  \thanks{ $^{\dagger}$\texttt{d.wang@northeastern.edu} Corresponding Author. Work done while Yunyao was at IBM Research.}\\ Northeastern University \\
}

\begin{document}
\maketitle
\begin{abstract}
Real-world domain experts (e.g., doctors) rarely annotate only a decision label in their day-to-day workflow without providing explanations.
Yet, existing low-resource learning techniques, such as Active Learning (AL), that aim to support human annotators mostly focus on the \textbf{label} while neglecting the \textbf{natural language explanation} of a data point.
This work proposes a novel AL architecture to support experts' real-world need for label and explanation annotations in low-resource scenarios.
Our AL architecture leverages \textbf{an explanation-generation model} to produce explanations guided by human explanations, \textbf{a prediction model} that utilizes generated explanations toward prediction faithfully, and
\textbf{a novel data diversity-based AL sampling strategy} that benefits from the explanation annotations. 
Automated and human evaluations demonstrate the effectiveness of incorporating explanations into AL sampling and the improved human annotation efficiency and trustworthiness with our AL architecture.
Additional ablation studies illustrate the potential of our AL architecture for transfer learning, generalizability, and integration with large language models (LLMs). 
While LLMs exhibit exceptional explanation-generation capabilities for relatively simple tasks, their effectiveness in complex real-world tasks warrants further in-depth study.

\end{abstract}

\section{Introduction}

\begin{figure}[!t]
    \centering
    \includegraphics[clip,width=.95\columnwidth]{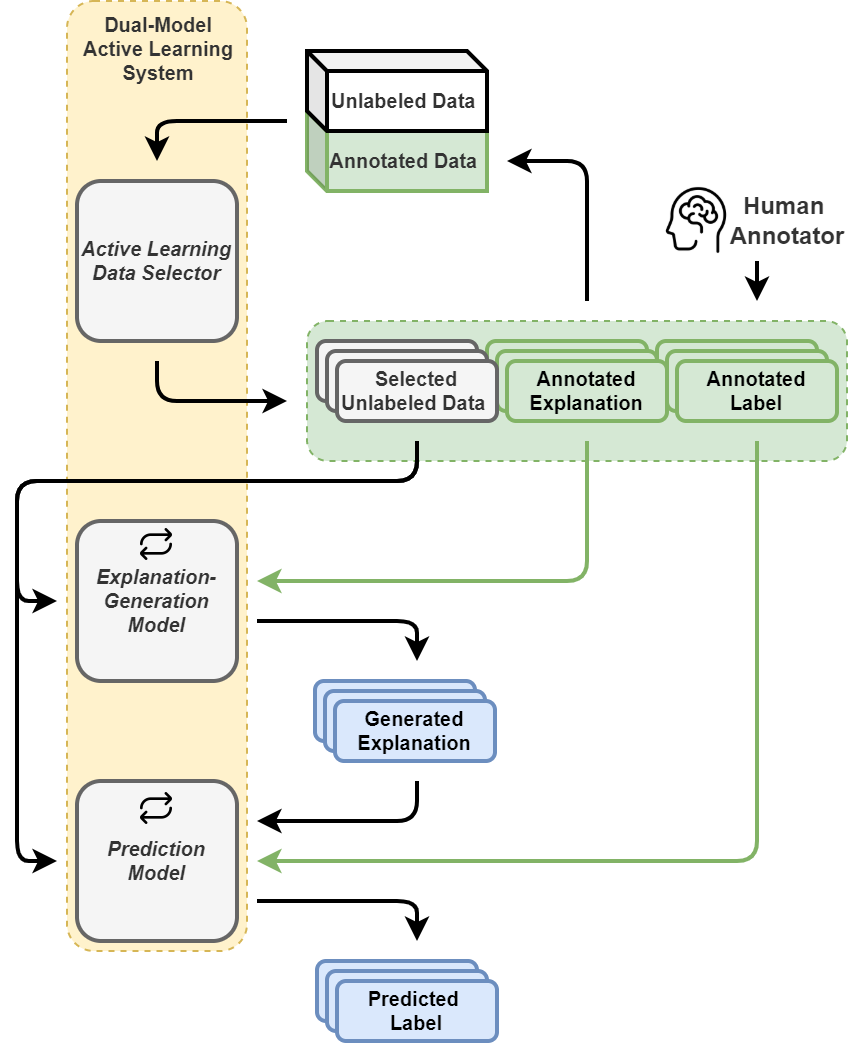}%

    \caption{Our dual-model AL system architecture at every iteration: 1) the AL data selector chooses a few unlabeled examples; 2) human annotators provide an explanation and label for each data instance; 3) the annotated explanations are used to finetune the explanation-generation model; 4) the annotated labels and generated explanations are used to finetune the prediction model. Then, humans can review the predicted labels and generated explanations for unlabeled data and start the next iteration. \textcolor{cexp}{Green arrows} indicate the training target.}
    
    \vspace{-1em}
    \label{fig:system_arch}
\end{figure}


State-of-the-art (SoTA) language models \cite{devlin-etal-2019-bert, radford2019language, winata-etal-2021-language} demonstrate astonishing performance on various NLP tasks, including Question Answering (QA) and Question Generation (QG)~\cite{rajpurkar-etal-2016-squad, duan-etal-2017-question, kocisky-etal-2018-narrativeqa, yao-etal-2022-ais}, 
Natural Language Inference (NLI)~\cite{bowman-etal-2015-large, wang-etal-2018-glue}, etc. 
Despite the superior generative capabilities, the lack of faithful explainability within these ``black boxes'' may lead to mistrust of their predictions~\cite{lipton2018mythos}, where humans, on the other hand, can develop intermediate rationales to facilitate the decision-making process.

The lack of explainability and untrustworthiness of models is magnified in the real world~\cite{drozdal2020trust}, where domain experts rarely only annotate a decision label in their daily workflow without providing explanations (i.e., clinical diagnoses by clinicians)~\cite{zhang2023rethinking}, and humans need explanations to understand and trust model predictions~\cite{zhang2021effect}. 
Therefore, a few approaches were proposed to retrospectively analyze the probability distribution within the model or ask models to generate explanations along with predictions~\citep{ribeiro2016should, lundberg2017unified, yu-etal-2019-rethinking, rajagopal-etal-2021-selfexplain, chen2021generate}, despite, the former is still very difficult for laymen to understand while the latter explanations are not faithful toward predictions.

As researchers looked into the quality~\cite{carton-etal-2020-evaluating, yao2023human} of human-annotated natural language explanations~\cite{camburu2018snli, rajani-etal-2019-explain, aggarwal-etal-2021-explanations}, they discovered numerous issues in existing datasets~\citep{geva2019we, chmielewski2020mturk, narang2020wt5, sun2022investigating}, such that the human annotations are of low quality and significant inconsistency. 
Furthermore, the ever-increasing costs in terms of labor, finances, and time for large-scale, high-quality data annotations remain a persistent challenge for the research community.
This challenge has given rise to various methodologies to reduce reliance on human annotations, such as Active Learning (AL)~\cite{settles2009active}.
AL is a human-in-the-loop framework that utilizes AL sampling strategies to iteratively select a small number of representative examples, request oracle annotations, and subsequently fine-tune the model using the annotated data. 
However, prior AL works predominantly focus on labels and overlook the fact that real-world scenarios often need both labels and natural language explanations.



In this work, we propose a dual-model AL architecture for human annotation of labels and explanations, drawing inspiration from the human decision-making process. Our system consists of: \\
\indent 1) An explanation-generation model guided by human-provided explanations \\
\indent 2) A prediction model that accepts the data content and the generated explanations for prediction. 

We integrate AL to reduce human annotation efforts and establish human trustworthiness by actively engaging humans in the training process.
We design a novel data diversity-based AL sampling strategy to select the most representative examples by exploiting the explanation annotations, which is analogous to the prevalent core-set~\cite{sener2017active} strategy.
Our AL architecture aims to support low-resource model predictions and AI trustworthiness by explicitly generating natural language explanations.
Specifically, we request label and free-form explanation annotations for a very limited number of examples (e.g., $3$ or $10$) selected by our AL sampling strategy at every AL iteration. Subsequently, the generated explanations serve as input for the final prediction, demonstrating the potential for these explanations to support the model's predictions faithfully.


We conduct two AL simulations with different amounts of samplings and iterations on a large-scale NLI dataset with human-annotated explanations to justify incorporating explanations in AL data selection can consistently outperform random, traditional data diversity-based, and model probability-based sampling strategies.  %
We make the code publically available\footnote{\url{https://github.com/neuhai/explanation-enriched-active-learning}}. 

A human evaluation of perceived validity, explainability, and preference of the generated explanations among our system, a SoTA explanation-generation system, and human-annotated explanations shows that, despite human explanations being ranked highest, explanations generated by our system are preferred over the SOTA system.
Additionally, we conduct three ablation studies to explore the capability and potential of our proposed AL architecture in transfer learning, generalizability, and incorporating large language models (LLMs) for explanation generation to further reduce human efforts.
LLMs demonstrate exceptional explanation-generation capabilities on relatively simple tasks. However, their effectiveness in handling complex real-world tasks warrants in-depth study.



\section{Related Work}

\subsection{Datasets with Natural Language Explanations}
\citet{wiegreffe2021teach} conducted a comprehensive review of 65 datasets with explanations and provided a 3-class taxonomy: highlights, free-text, and structured. Among the large-scale datasets with free-text explanations, \textbf{e-SNLI}~\citep{camburu2018snli} is a prominent one, which extended the Stanford Natural Language Inference (SNLI) corpus~\citep{bowman-etal-2015-large}, a classification task to determine the inference relation between two textual contexts (premise and hypothesis): entailment, contradiction, or neutral. 
The e-SNLI dataset (examples are shown in Appendix~\ref{sec:esnli}) contains human-annotated free-form explanations for $549,367$ examples in train, $9,842$ in validation, and $9,824$ in test split.

Another popular group of datasets extended the Commonsense QA (CQA v1.0 and v1.11 versions) datasets \citep{talmor-etal-2019-commonsenseqa}, including two variants of Cos-E dataset (\textbf{CoS-E v1.0} and \textbf{CoS-E v1.11}~\citep{rajani-etal-2019-explain}) and the ECQA~\cite{aggarwal-etal-2021-explanations} dataset. 
Many recent works~\citep{narang2020wt5, sun2022investigating} have found explanations in CoS-E to be noisy and low-quality, and thus, \citet{aggarwal-etal-2021-explanations} carefully designed and followed the explanation annotation protocols to created \textbf{ECQA}, which is of higher quality compared with CoS-E. 

In this paper, we leverage the e-SNLI dataset as the benchmark dataset for our AL simulation experiment because 1) the classification task is popular and representative, 2) the massive data size ensures data diversity, and 3) explanations for a classification task may provide more effective help compared to CQA task where training and testing data may be unrelated.  
We additionally conduct an ablation study on the ECQA dataset to explore the generalizability of our proposed AL architecture.

\subsection{Active Learning for Data Annotation}
Owning to the paucity of high-quality, large-scale benchmarks for a long tail of NLP tasks, learning better methods for low-resource learning is acquiring more attention, such as Active Learning (AL)~\citep{sharma-etal-2015-active, shen-etal-2017-deep, ash2019deep, teso2019explanatory, kasai-etal-2019-low, zhang-etal-2022-allsh}. 
AL iteratively 1) selects samples from the unlabeled data pool (based on AL sampling strategies) and queries their annotation from human annotators, 2) fine-tunes the underlying model with newly annotated data, and 3) evaluates model performance. 


A few AL surveys~\cite{settles2009active, olsson2009literature, fu2013survey, schroder2020survey, ren2021survey} of sampling strategies provide two high-level selection concepts: data diversity and model probability. 
We propose a novel \textbf{data diversity-based} strategy that leverages human-annotated explanations to select data. Our data selector shares a similar concept with the established data-based clustering strategies~\citep{xu2003representative, nguyen2004active} and core-set~\cite{sener2017active} that aim to select the most representative data while maximizing diversity. 
Compared with model probability-based strategies, data diversity-based ones are model-agnostic and need much less computing resources, whereas the former requires inference on unlabeled examples to calculate probability.  

In addition to AL, \citet{marasovic-etal-2022-shot} introduces a few-shot self-rationalization setting that asks a model to generate free-form explanations and the labels simultaneously. 
Similarly, \citet{bhat-etal-2021-self} proposes a multi-task self-teaching framework with only 100 train data per category, and 
\citet{bragg2021flex} provides guidance on unifying evaluation for few-shot settings. 

\subsection{Natural Language Explanation Generation}

Different approaches have been explored to enhance the model's explainability by asking them to generate natural language explanations. 
Some of them~\citep{talmor2020leap, tafjord-etal-2021-proofwriter, latcinnik2020explaining} propose systems to generate text explanations for specific tasks.  
\citet{dalvi2022towards} propose a 3-fold reasoning system that generates a reasoning chain and asks users for correction. 
Other recent works~\cite{paranjape-etal-2021-prompting, liu-etal-2022-generated, chen-etal-2022-rationalization} explore different prompt-based approaches to generate additional information for the task and examine the robustness and validity. 
We believe that our dual-model system provides and uses explanations explicitly towards prediction, while the self-rationalization setting falls short. \citet{hase-bansal-2022-models} argues that explanations are most suitable as input for predicting, and \citet{kumar-talukdar-2020-nile} designed a system to generate label-wise explanations, which is aligned with our design hypothesis. Nevertheless, there exist other works~\citep{wiegreffe-etal-2021-measuring, marasovic-etal-2022-shot, zelikman2022star} 
that explore the use of self-rationalization setting. We include the self-rationalization setting in our human evaluation of the explanation quality in Section~\ref{sec:human_eval}.

\section{Dual-Model AL System}

\subsection{System Architecture}
\label{sec:system_arch}

Figure~\ref{fig:system_arch} illustrates our proposed dual-model AL framework. 
The system comprises three primary modules: 1) \textbf{an explanation-generation model} that takes the data, fine-tunes on human-annotated explanations, and generates free-form explanations; 2) \textbf{a prediction model} that accepts the data content and the generated explanations as input, fine-tunes on human-provided labels, and predicts the final label; 3) \textbf{an AL data selector} that selects a set of representative examples in terms of the semantic similarity between each unlabeled data text and labeled data's human explanations. 
The AL data selector plays a crucial role in finding a small, highly representative set of samples at every iteration, and further details of our AL selector are in Section~\ref{aldataselector}.

In each AL iteration, after the data selector samples unlabeled examples for human annotations, we first fine-tune the explanation-generation model supervised by human-provided free-form explanations.
Then, we instruct this model to generate explanations for the same set of data.
Subsequently, we fine-tune a prediction model using the data content and explanations generated by the previous model as input, supervised by human-annotated labels. 
The fine-tuning process teaches the prediction model to rely on the explanations for predictions~\cite{yao2023human}. 
Additionally, we fine-tune the prediction model with model-generated explanations instead of human-annotated ones for better alignment during inference, especially when no human annotations are available. 
After each AL iteration, we evaluate the framework on a standalone evaluation data split.

\begin{table}[!t]
\centering
\small
\begin{tabular}{p{.45\textwidth}}
    \toprule
    \textbf{Explanation-generation Model:} \\
    \textcolor{cpre}{\textbf{Training Input }} \textbf{explain: }what is the relationship between \textit{[hypothesis]} and \textit{[premise]} \textbf{choice1:} entailment \textbf{choice2:} neutral \textbf{choice3:} contradiction \\
    \textcolor{clabel}{\textbf{Training Target }} \textit{[human annotated explanations]} \\
    \textcolor{cexp}{\textbf{Model Generation }} \textit{[generated free-form explanation]} \\
    \midrule
    \midrule
    \textbf{Prediction Model:} \\
    \textcolor{cpre}{\textbf{Training Input }} \textbf{question:} what is the relationship between \textit{[hypothesis]} and \textit{[premise]} \textbf{choice1:} entailment \textbf{choice2:} neutral \textbf{choice3:} contradiction \textbf{<sep> because} \textit{[generated free-form explanation]}\\
    \textcolor{clabel}{\textbf{Training Target }} \textit{[human annotated label]} \\
    \textcolor{cexp}{\textbf{Model Prediction }} \textit{[predicted category]} \\
    \bottomrule
    
\end{tabular}
\vspace{-0.5em}
\caption{ The prompt-based input templates for both models in our system, with the e-SNLI~\citep{camburu2018snli} dataset as an example. }
\label{tab:data_format}
\vspace{-1.5em}
\end{table}

Both the explanation-generation model and the prediction model can be any SoTA sequence-to-sequence models, such as BART~\citep{lewis-etal-2020-bart} and T5~\citep{raffel2020exploring}. In this work, we utilize T5 as the backbone for both models and design a prompt-based input template for both models, as shown in Table~\ref{tab:data_format}, inspired by a few existing works~\cite{schick-schutze-2021-exploiting, gao-etal-2021-making, zhou-etal-2023-flame}. 
To elucidate how each prompt addresses a different part of data content: \\
\indent 1) ``\textit{explain:}'' and ``\textit{question:}'' are the leading prompts in the explanation-generation model and the prediction model, respectively, indicating different tasks for both models and are followed by the original task content. 
For the e-SNLI dataset, the task content becomes ``what is the relationship between'' the hypothesis and premise sentences; \\
\indent 2) ``\textit{choice$N$}'' is followed by candidate answers, where $N\in[1,3]$ for the e-SNLI dataset corresponds to entailment, neutral, and contradiction. 
We pass the choices to the explanation-generation model, expecting that it will learn to generate free-text explanations that may reflect potential relationships between the data content and the task; \\
\indent 3) for the prediction model, an additional prompt ``\textit{because:}'' is followed by the explanations generated by the explanation-generation model. We use a special token to separate the original task content and the explanation."

\subsection{AL Data Selector}
\label{aldataselector}

\begin{algorithm}
\small
\caption{Our Data Diversity-based AL Selector}
\label{alg:selector}
\textbf{Variables:} 
\begin{algorithmic}
\STATE $D_{train}\Rightarrow$ unlabeled data in train split 
\STATE $D_{prev}\Rightarrow$ previously-annotated data 
\STATE $d_{p}^{data}\Rightarrow$ data content as a string of $d_{p}$ (for e-SNLI, it is the premise and hypothesis
\STATE $d_{p}^{exp}\Rightarrow$ previously-annotated free-form explanation of $d_{p}$
\STATE $x\Rightarrow$ number of data to be selected each iteration
\STATE $n_{train}=len(D_{train});\:n_{prev}=len(D_{prev})$

\FOR {$D_i \in D_{train}$}
    \IF {$iteration == 0$}
        \STATE $score_{d_i}=\frac{1}{n_{train}}\cdot\sum_{d_{p}\in D_{train}} similarity(d_{i}^{data},d_{p}^{data}) $
    \ELSE
        \STATE $score_{d_i}=\frac{1}{n_{prev}}\cdot\sum_{d_{p}\in D_{prev}} similarity(d_{i}^{data},d_{p}^{exp}) $
    \ENDIF
\ENDFOR
\STATE $D_{train}^{'}=rank\:D_{train}\:by\:score$
\STATE $D_{selected}=\:select\:x\:data\:from\:D_{train}^{'} with\:equal\:intervals$
\STATE \textbf{Human annotation on} $D_{selected}$
\STATE $D_{train}-=D_{selected}; D_{prev}+=D_{selected}$

\end{algorithmic}

\end{algorithm}
\vspace{-1em}


According to recent surveys of AL ~\cite{settles2009active, olsson2009literature, fu2013survey, schroder2020survey, ren2021survey},
there are two primary approaches for AL data selection: model probability-based and data diversity-based approaches. 
Model probability-based approaches, firstly, aim to select examples about which the models are least confident. 
These approaches involve conducting inference on unlabeled data at every iteration, which consumes more time and computing resources. Unlike data diversity-based approaches, they are not model-agnostic, which may affect the effectiveness of the sampling strategies depending on the model in use.

Secondly, data diversity-based approaches leverage various data features, such as data distribution and similarity, to select a representative set of examples from the candidate pool while maximizing diversity. 
This paper introduces a \textbf{data diversity-based} AL selection strategy that shares a concept similar to traditional data-based clustering strategy \citet{nguyen2004active} and core-set strategy. 
However, our strategy differs from traditional strategies because ours incorporates human-annotated explanations for selection. 
More specifically, our data selector aims to choose examples that are representative of the unlabeled data pool in terms of average similarity to human-annotated explanations of all previously-labeled data while maximizing the diversity of newly-selected data. 

We assume that \textbf{human-annotated explanations contribute significantly to the model's prediction and convey more information than the original data content alone}.
These explanations can reveal underlying relationships between concepts in the data content and the relations between the data content and choices.
For instance, in the e-SNLI dataset, the data content consists of the concatenation of hypothesis and premise sentences. 
Later, we construct a baseline selector in the AL simulation experiment (Sec.~\ref{sec:als}) with the same setup, except that it only compares the similarity between data content.
Additionally, we include random baseline and probability-based baseline strategies.
Our results demonstrate that using human-annotated explanations for data selection consistently leads to improved prediction performance compared to using data content alone.

\begin{figure}[!t]
    \centering
    \includegraphics[clip,width=.95\columnwidth]{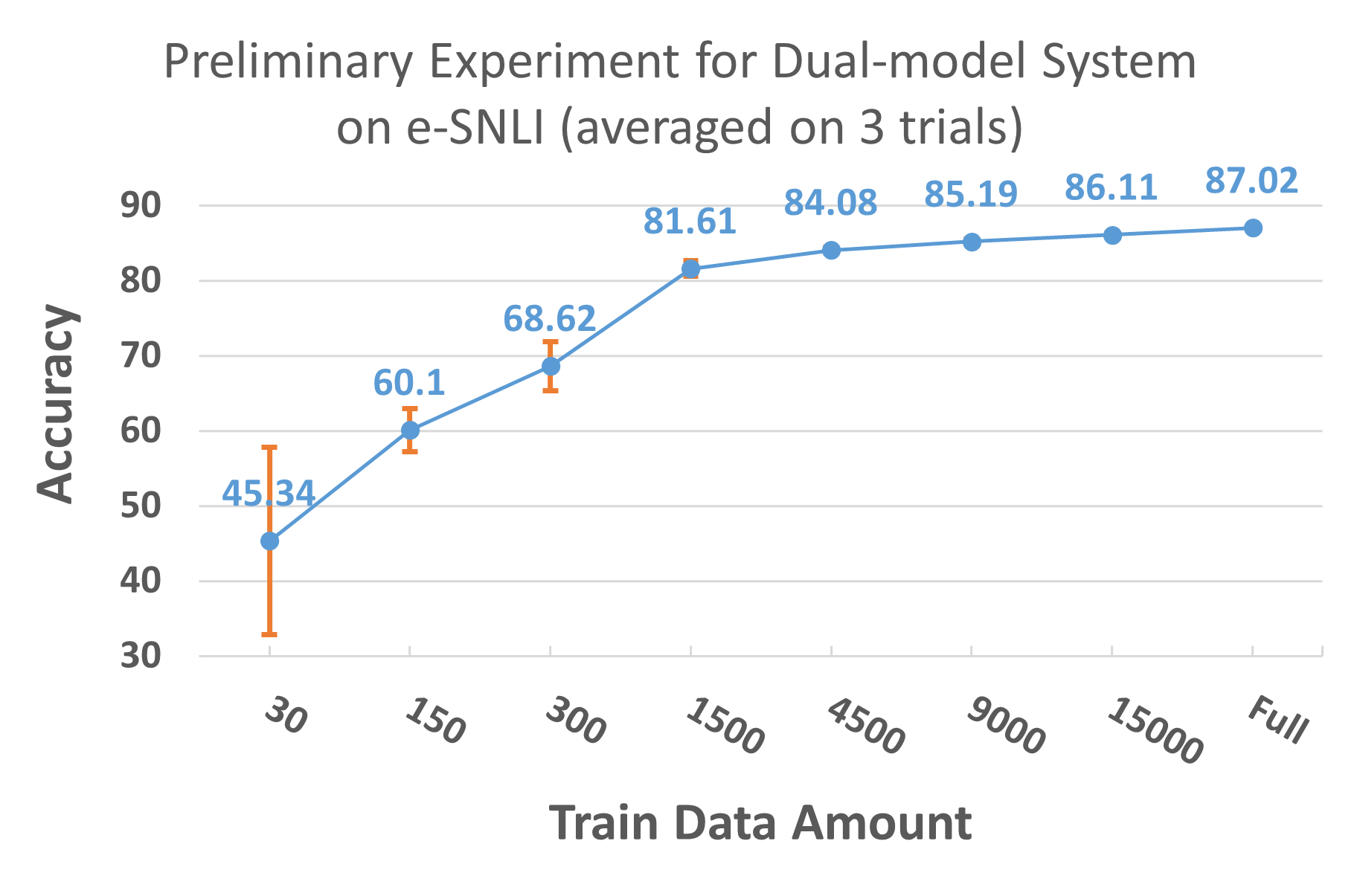}%
    \vspace{-0.5em}
    \caption{ Preliminary experiment result of our dual-model system on e-SNLI~\citep{camburu2018snli} dataset. }
    
    \vspace{-1em}
    \label{fig:pe_result}
\end{figure}

Here we delve into the details of our data-based AL data selector (shown in Algorithm~\ref{alg:selector}). 
For each unlabeled data instance, we use sentence-transformers~\cite{reimers-2019-sentence-bert} to calculate the semantic similarity between its data content and every previously annotated explanation. 
Then, we take the averaged similarity scores for each unlabeled example and rank all the unlabeled data in terms of the average similarity score. 
To select the most representative data in the candidate pool while maximizing diversity, we choose examples from the ranked data list with equal intervals. 
Note that in the first iteration, since no previously annotated explanations are available, we compare the similarity between the data content.



\section{Evaluation}

We conduct the AL simulation experiment with the e-SNLI~\citep{camburu2018snli} dataset. 
The primary objective is to justify that
our proposed dual-model framework, when combined with human-annotated explanations in AL data selection, can effectively identify more representative and helpful data from a reasonably large-scale dataset.

Given that e-SNLI dataset comprises a substantial $549,367$ examples in the train split, we performed a preliminary experiment to determine a reasonable number of candidate data for the AL simulation. 
This approach aims to save time and computing resources.
Our goal is to identify an ideal candidate data size that would not introduce potentially biased feature distributions or significantly degrade model performance when compared to fine-tuning on the full dataset.
We employ the pre-trained T5-base~\citep{raffel2020exploring} as the backbone for all the experiments and provide the hyperparameters in Appendix~\ref{app:hyperparameters}.


\subsection{Preliminary Experiment}
\label{sec:pe}

The expected outcome of the aforementioned preliminary experiment is
1) to determine the upper bound of performance and observe how the performance of our dual-model system gradually decreases as we reduce the amount of training data, and 
2) to identify a suitable candidate data size for the AL simulation. 

We also randomly sample the same amount of data for each category in the preliminary experiment to minimize potential bias introduced by uneven distribution, especially when the sampling size per iteration is very small.
Specifically, we select eight different sampling amounts per category from the e-SNLI training split, ranging from [$10$, $50$, $100$, $500$, $1500$, $3000$, $5000$] and the complete data per category. Since the e-SNLI dataset consists of three categories: entailment, neutral, and contradiction, the total sampling size in each setting becomes [$30$, $150$, $300$, $1500$, $4500$, $9000$, $15000$, and $549,367$ (full train split)], respectively, as shown in Figure~\ref{fig:pe_result}.

For each sampling setting, we conduct three trials to obtain an averaged result. In each trial, we fine-tune the explanation generation model and the prediction model once and conduct a hyperparameter search.
The framework is then evaluated on the test split of e-SNLI ($9,824$ examples). 

The preliminary experiment results are shown in Figure~\ref{fig:pe_result}, where the \textcolor{blue}{blue dot} denotes the averaged prediction accuracy (in percentages) at each setting, and the \textcolor{red}{red bar} indicates the standard deviation of accuracy among three trials. 
Notably, with more than $1,500$ data per category, the performance drop compared to the full train split is inconspicuous ($84.08\%$ to $87.02\%$), while the standard deviation is below 0.5\%.
This observation indicates that using $1\%$ of the original training data size only leads to a performance drop of merely $3\%$. Additionally, we found that even with only 10 data points per category (30 data in total), our system still achieves an average accuracy of 45\%, although the deviation is relatively significant. 
Furthermore, when we extend the training data size from $100$ to $500$ data per category ($300$ to $1500$ in total), a reasonably applicable setting in real-world scenarios, the accuracy can reach over $80\%$ accuracy, showing promising results considering that the amount of training data is much smaller than the size of evaluation samples.


\subsection{Simulation Experiment: Evaluation Setup }
\label{sec:als}

Based on the findings from the preliminary experiment, we decide to use $3,000$ examples per category ($9,000$ in total) as the candidate unlabeled data pool for the Active Learning simulation. 

Inspired by the few-shot evaluation guidance~\cite{bragg2021flex}, we conduct \textbf{80} trials for each AL setting and calculate the averaged performance for ours and the baseline data selectors at every iteration. 
During each trial, we start by randomly selecting $3,000$ examples per category from the complete train dataset, then use the same data to conduct AL simulations with different data selectors in our dual-model framework. 
This way, we can ensure the performance differences during each trial are not due to different unlabeled data pools but to actual differences in the performance of the AL data selectors. For the evaluation, we randomly sample 300 examples per category (900 in total) from the test split of e-SNLI every trial and evaluate with the same test data after each iteration. 

The AL simulation comprises two settings, where we simulate annotating 180 and 450 data instances, respectively. These two levels of data annotations reasonably mimic real-world scenarios where users have limited budgets, annotators, and data for annotation. Specifically, we experiment with the following two settings: 


\indent 1) For every iteration, select \textbf{3} examples per category (9 in total) with \textbf{20} iterations, which results in \textbf{180} examples altogether; \\
\indent 2) For every iteration, select \textbf{10} examples per category (30 in total) with \textbf{15} iterations, which results in \textbf{450} examples altogether. 



\begin{figure}[!t]
    \centering
    \subfloat[Setting 1: 9 examples per iteration + 20 iterations]{%
        \includegraphics[clip,width=.95\columnwidth]{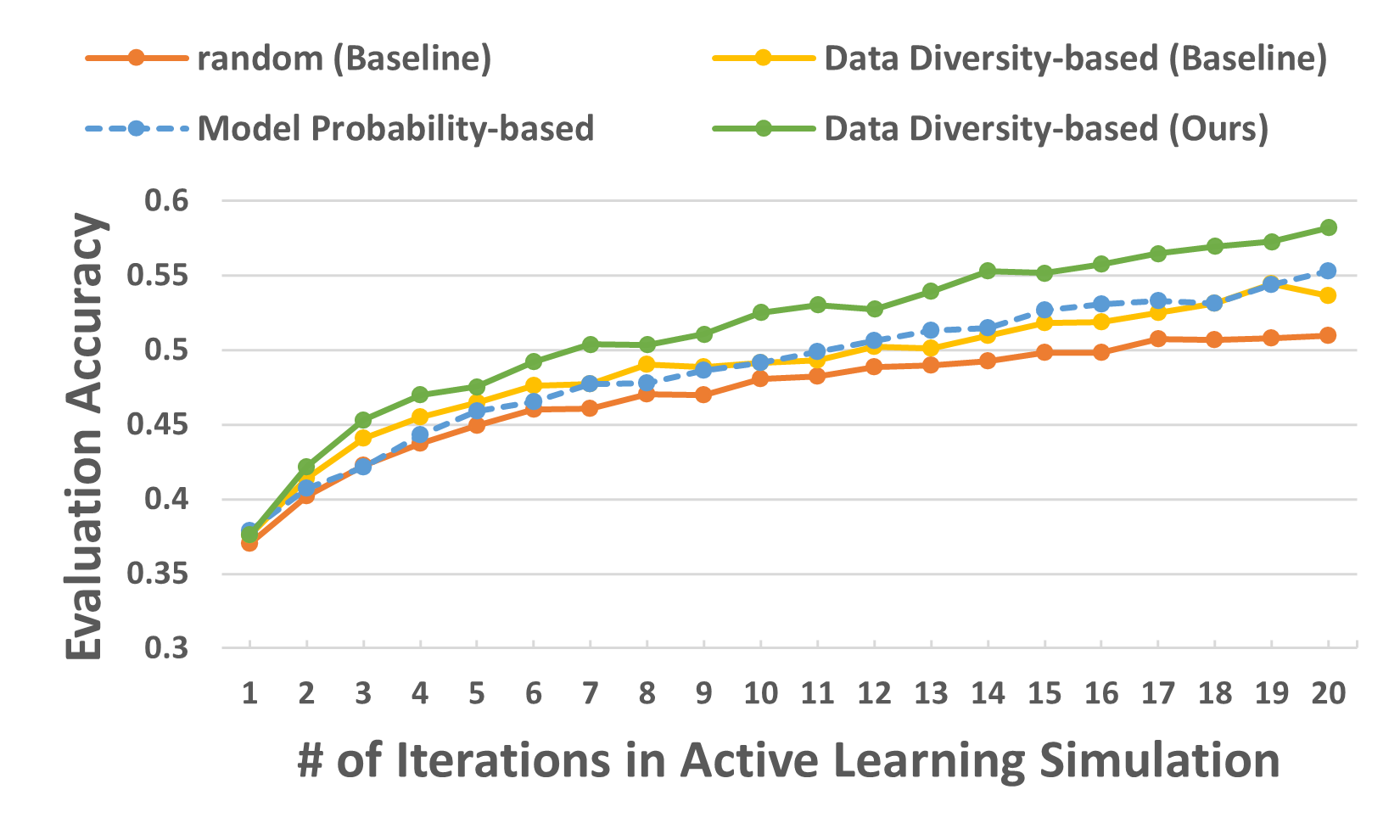}%
    }
    \hfill
    \subfloat[Setting 2: 30 examples per iteration + 15 iterations]{%
        \includegraphics[clip,width=.95\columnwidth]{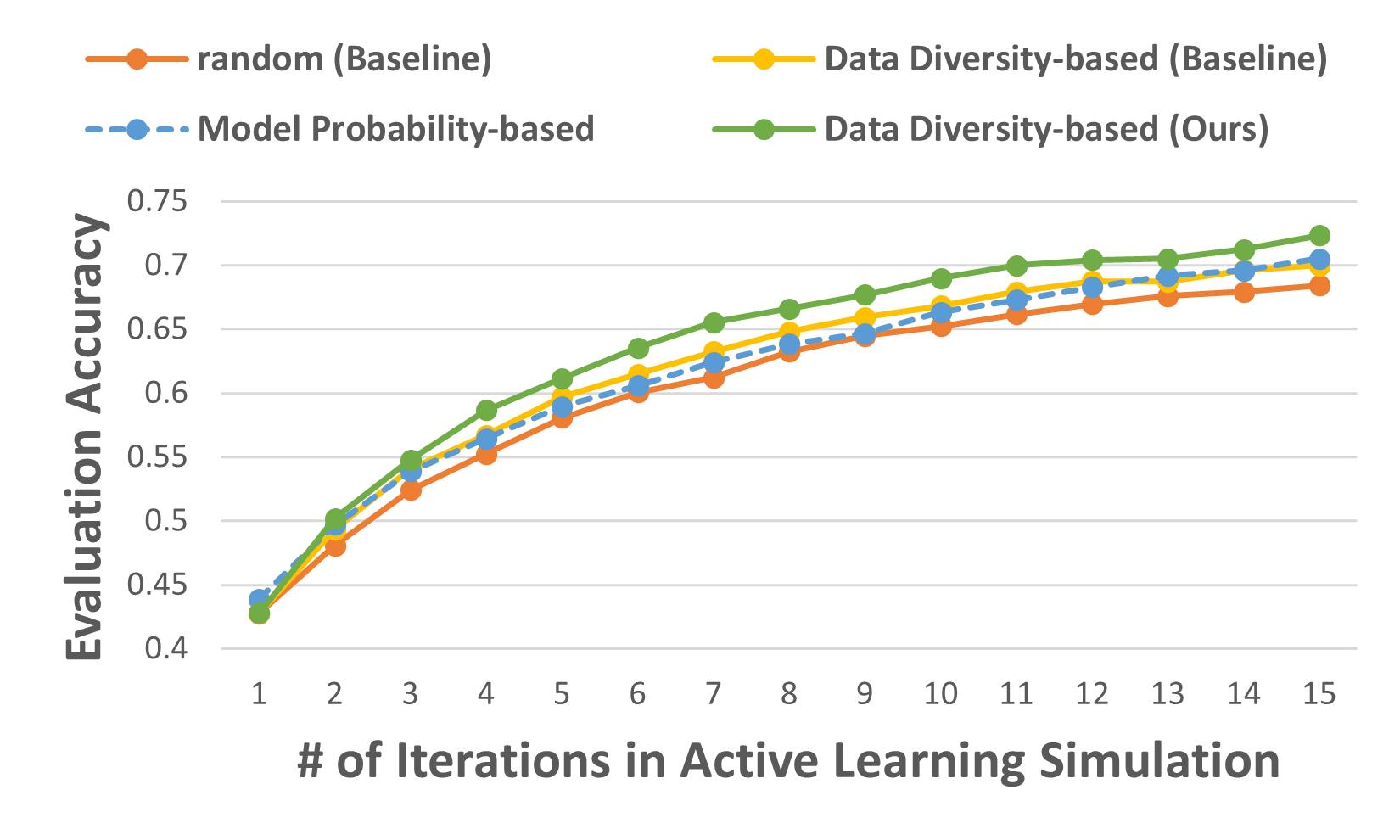}%
    }
    \vspace{-0.5em}
    \caption{ Results of AL Simulation experiment on our Dual-model system with different data selectors. }
    
    \vspace{-1em}
    \label{fig:al_result}
\end{figure}

Our AL simulation experiment involves our data selector, two baselines, and an additional model probability-based selector.
Our data selector, described in Section~\ref{aldataselector}, is a novel data diversity-based sampling strategy that leverages human-annotated explanations. 
For comparison, we use a random data selector as the basic benchmark and another traditional data diversity-based algorithm that shares the same procedures with ours, except that it only compares the similarity between each unlabeled data's content and the previously-labeled data's content, not using the human-annotated explanations. 
The probability-based selector conducts inference on unlabeled data and selects examples with the least probability at every iteration. 
We fix the same set of hyperparameters (Appendix~\ref{app:hyperparameters}). 

Worth noting that our data selector does not use task content in previously labeled examples; instead, we exclusively rely on human-annotated explanations to demonstrate their greater utility compared to task content.
In the first iteration, both ours and the data diversity baseline perform identically because no previously annotated data is available.

\subsection{Simulation Experiment: Result}
\label{sec:als_result}

The AL simulation results are presented in 
Figure~\ref{fig:al_result}.
To explain the diagrams in detail, each dot is the average accuracy on 80 trials at every iteration for each data selector. The \textcolor{green}{green}/\textcolor{yellow}{yellow}/\textcolor{red}{red}/\textcolor{blue}{blue} dots denote our data selector/data diversity-based baseline/random selector/model probability-based selector, respectively. 
We observe that 
our data selector consistently maintains an advantage over the traditional data-based sampling baseline, while the traditional one consistently beats the random baseline by a significant margin. Additionally, we observe that the model probability-based selector outperforms the random baseline in both settings.

To summarize,
our data selector outperforms both baselines in every iteration for both AL settings, indicating that 
\textbf{using human-annotated explanations in the data selector with our dual-model AL framework is more beneficial than using the data content alone.}
Even with only 180 and 450 data to be annotated in each setting, our system can achieve $55\%$ and $72\%$ accuracy on average, respectively. 
We anticipate that our experiment will reach a similar performance around 85\% as shown in Figure~\ref{fig:pe_result} but converge much faster than the random selector if we continue the AL process.

\subsection{Human Evaluation Setup and Results}
\label{sec:human_eval}

To qualitatively evaluate the explainability of the generated explanations from our system against a SoTA few-shot explanation-generation system, the self-rationalization baseline~\citep{marasovic-etal-2022-shot}, and the human ground-truth, we recruited three human participants to conduct a human evaluation following the prior literature~\cite{xu2022fantastic}.
The self-rationalization baseline is a T5-base model, which uses the same input template of our explanation-generation model shown in Table~\ref{tab:data_format} but asks the model to generate both the label and explanation simultaneously.

We leverage AL setting $1$ described in Section~\ref{sec:als} to fine-tune our system with a total of 180 examples over 20 epochs and use the same 180 examples to fine-tune the self-rationalization baseline. Both systems are used to infer the complete test split of e-SNLI after fine-tuning; then, we randomly sample 80 examples for the human study. 

For each data instance, the rater is presented with the textual content of the \textit{premise} and \textit{hypothesis} of the original data paired with three sets of \textit{labels} and \textit{explanations} from our system, baseline system, and the human-annotated ground-truth from the e-SNLI dataset. 
Participants who are not aware of the source of each label-explanation pair are asked to answer four questions with [Yes/No]:\\
\begin{small}
\indent 1) Is the \textcolor{clabel}{Prediction} correct? \\
\indent 2) Is the \textcolor{cexp}{Explanation} itself a correct statement? \\
\indent 3) Regardless of whether the AI Prediction and Explanation is correct or not, can the \textcolor{cexp}{Explanation} help you to understand why AI has such \textcolor{clabel}{Prediction}? \\
\indent 4) Will you trust $\&$ use this AI in real-world decision-making? \\
\end{small}
\indent To ensure inter-coder consistency, we first conduct a 30-min tutorial session to educate all three participants with ten examples to build a consensus among them. 
In the actual experiment, each of the three participants is then asked to rate 30 data instances (20 unique ones and 10 shared ones), which make up a total of 70 data instances, and 360 ratings (3 rater*30 instances*4 questions). We first calculated the Inter-Rater Reliability score (IRR) among them for each of the four questions. With the IRR score of (Q1: 1, Q2: 0.89, Q3: 0.98, Q4: 0.87), we are confident that the three coders have the same criteria for further result analysis. 


\begin{table}[t]
\centering
\resizebox{.95\columnwidth}{!}{%
\small
\begin{tabular}{lcccc}
\toprule

\textbf{Yes / No Count} & \textbf{Label} &   \textbf{Exp.} &   \textbf{Exp. $\rightarrow$ Label} &   \textbf{Trustworthy AI} \\

\cmidrule(lr){1-1} \cmidrule(lr){2-2} \cmidrule(lr){3-3} \cmidrule(lr){4-4} \cmidrule(lr){5-5} 
Ground-truth    &\hfil   83 / 7  &\hfil   86 / 4  &\hfil   87 / 3  &\hfil   78 / 12 \\
Dual-model (ours)  &\hfil  64 / 26 &\hfil   68 / 22 &\hfil   48 / 42 &\hfil   35 / 55 \\
Self-rationalization &\hfil   42 / 48 &\hfil   67 / 23 &\hfil   51 / 39 &\hfil   21 / 69 \\

\bottomrule
\end{tabular}
}

\caption{ Human evaluation results.  }
\vspace{-1em}
\label{tab:he_1}
\end{table}

Our questions all have binary responses, and we rely on Chi-square analysis~\citep{elliott2007statistical} to examine the statistical significance of the rating groups' differences.
As shown in Table~\ref{tab:he_1}, the participants rated human ground-truth explanations highest across all four dimensions.
Between our system and the few-shot self-rationalization system (baseline), participants believe our systems' predicted labels are more likely to be correct, with 64 `valid' ratings out of 90 for our system versus 42 out of 90 ratings for the baseline. Chi-square test indicates such a difference is statistically significant ($\chi^2(1)=21.61, p<0.01$).

When asked whether they would trust the AI if there were such AI systems supporting their real-world decision-making, 35 out of 90 answered `Yes' for our system, and it is  significantly better than the baseline system 
(21 `Yes' out of 90) 
($\chi^2(1)=12.17, p<0.01$). 
In comparison, 78 out of 90 times people voted that they would trust the human-annotated explanation's quality.

As for Question 2 (``the validity of the generated explanation'') and Question 3 (``whether the generated explanation is supporting its prediction''), the human evaluation fails to suggest statistically meaningful results between our system and the baseline system ($\chi^2(1)=0.06, p=0.89$ for explanation validity, and $\chi^2(1)=0.41, p=0.52$ for explanation supporting prediction).
In summary, human participants believe our system can outperform the baseline system on the label prediction's quality and the trustworthiness of AI dimensions. Still, there is a large space to improve as human evaluators believe the ground-truth label and explanation quality is much better than either AI system.

\subsection{Ablation Study 1: Transfer to Multi-NLI}
\label{sec:abla_eval}

We conduct an ablation study with transfer learning through AL simulation from e-SNLI to Multi-NLI~\cite{williams-etal-2018-broad}. 
This study explores whether the explanation-generation model trained on e-SNLI is helpful for AL on a similar task.

The transfer-learning ablation study consists of the following steps: 
1) fine-tune an explanation-generation model using our AL framework on the e-SNLI dataset; 
2) freeze the explanation-generation model and use it to generate explanations in the AL simulation for Multi-NLI;
3) fine-tune the prediction model for Multi-NLI at every iteration. 
Unlike the e-SNLI experiment, our AL data selection algorithm will use model-generated explanations to select examples at every iteration in the transfer learning AL simulation. We fine-tune the explanation-generation models on e-SNLI with the same two settings in the previous experiment, average the result on 15 trials of experiments, and keep consistent with every other hyper-parameters.

The ablation results are shown in Figure~\ref{fig:al_tr_result} of Appendix~\ref{app:tr_results}. The \textcolor{blue}{blue}/\textcolor{red}{red} lines denote the explanation-generation model is fine-tuned on e-SNLI with each setting in Section~\ref{sec:als} correspondingly. 
We observe that the explanation-generation model consistently provided helpful explanations, leading to an improvement in the system's prediction performance, with accuracy reaching more than 65\%. 
In addition, the explanation-generation model fine-tuned on more data can perform better,  suggesting that it had learned to generate more helpful explanations.

\begin{figure}[!t]
    \centering
    \subfloat[Setting 1: 9 examples per iteration + 20 iterations]{%
        \includegraphics[clip,width=.95\columnwidth]{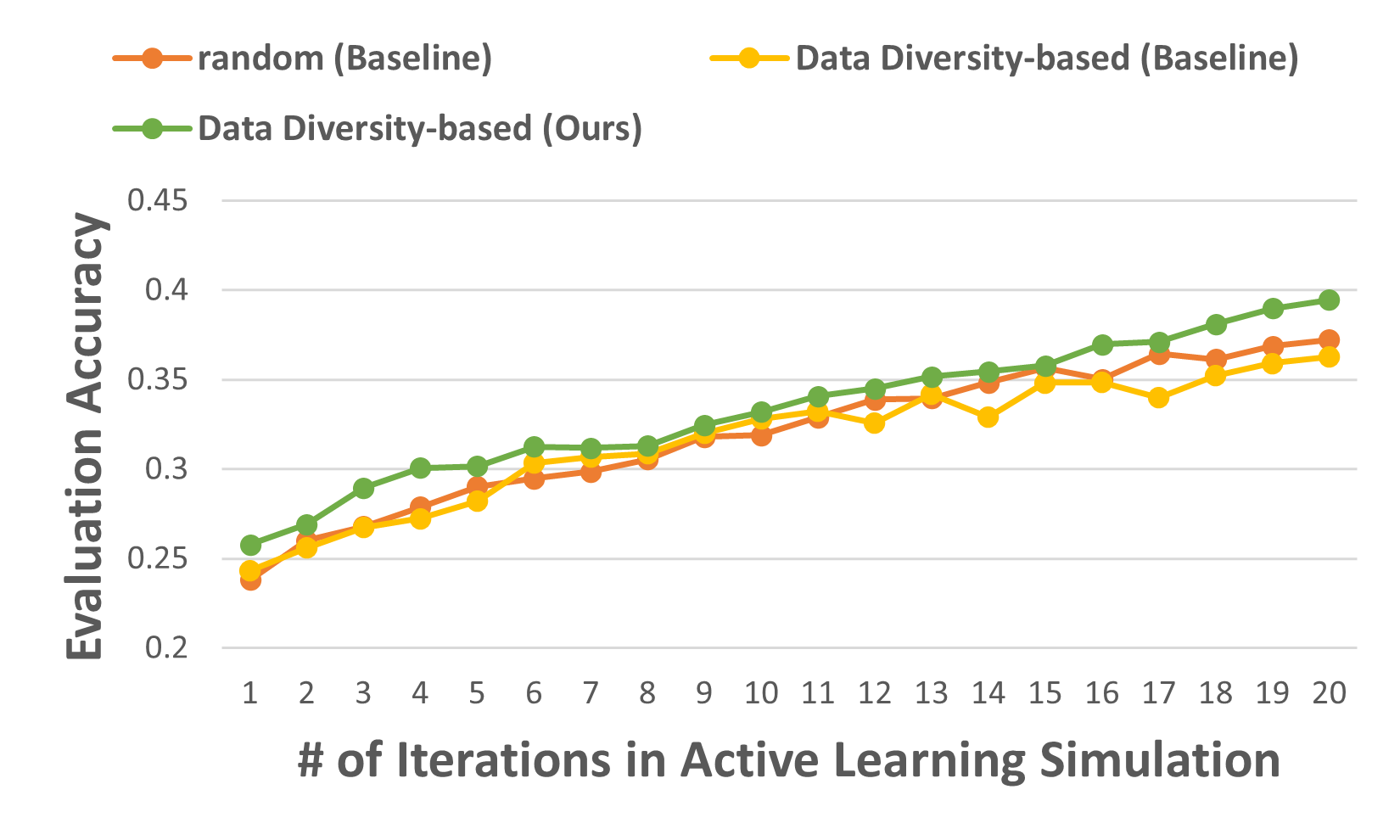}%
    }
    \hfill
    \subfloat[Setting 2: 30 examples per iteration + 15 iterations]{%
        \includegraphics[clip,width=.95\columnwidth]{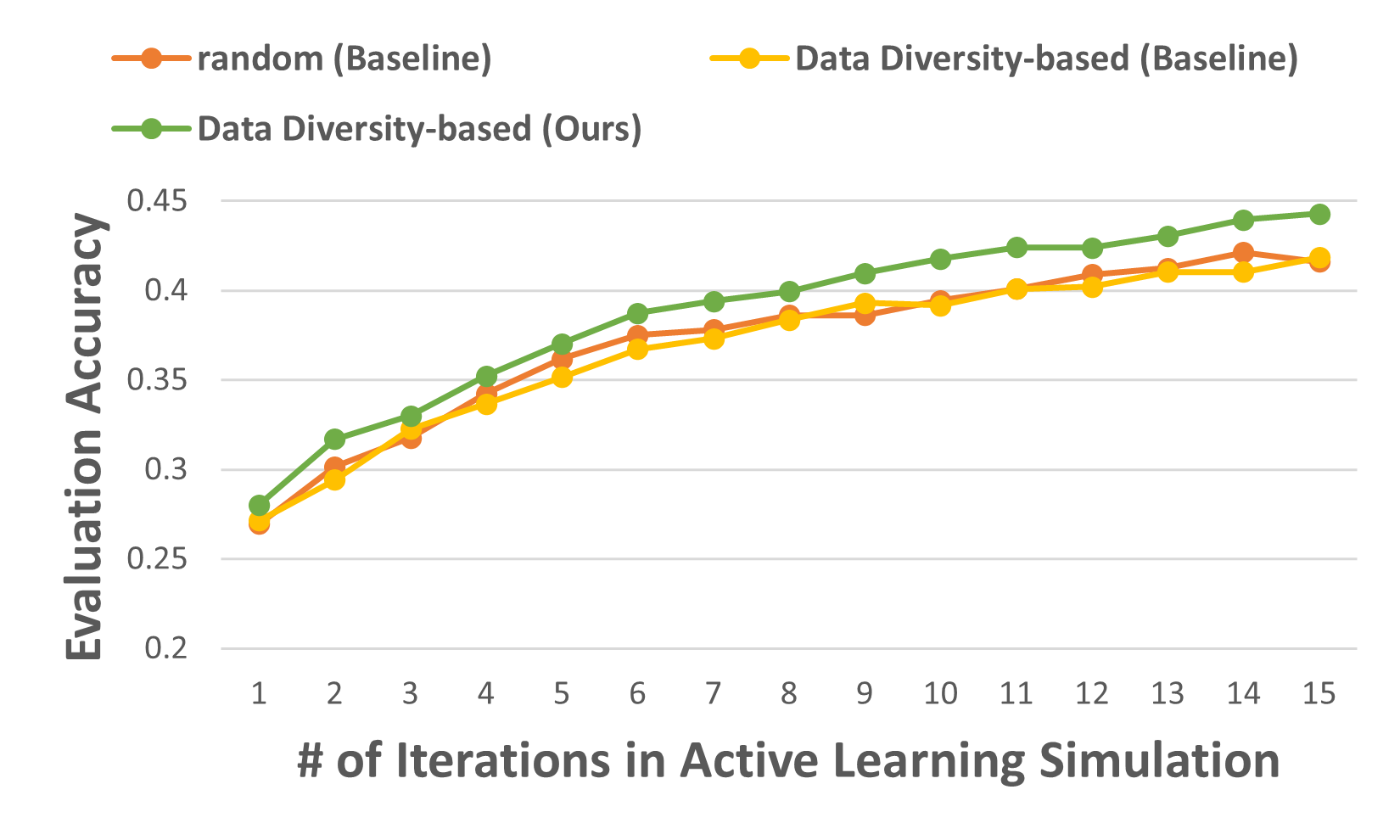}%
    }
    \caption{ Ablation study results of AL simulation experiment on our Dual-model system with different data selectors on ECQA dataset. }
    
    \vspace{-1em}
    \label{fig:al_result_ecqa}
\end{figure}

\subsection{Ablation Study 2: Our AL Framework on ECQA}
\label{sec:abla_ecqa}

We additionally conduct an AL Simulation experiment on ECQA~\cite{aggarwal-etal-2021-explanations} (a recent dataset extends the CommonsenseQA dataset with high-quality human-annotated explanations) with our data selector, random baseline, and similarity-based baseline that does not use explanations. 
We comply with the same experiment settings for the e-SNLI AL simulations described in Section~\ref{sec:als}.
The results are shown in Figure~\ref{fig:al_result_ecqa}, where our proposed data selection strategy can consistently outperform both baselines in both simulation settings. 
Interestingly, the similarity-based baseline performs similarly to the random baseline, which could be because using data content alone is not sufficient to select more helpful and representative examples while using human-annotated explanations can facilitate better data selection consistently.

\subsection{Ablation Study 3: LLM for Explanation Generation}
\label{sec:abla_llm}

\begin{figure}[!t]
    \centering
    \subfloat[e-SNLI Dataset]{%
        \includegraphics[clip,width=.95\columnwidth]{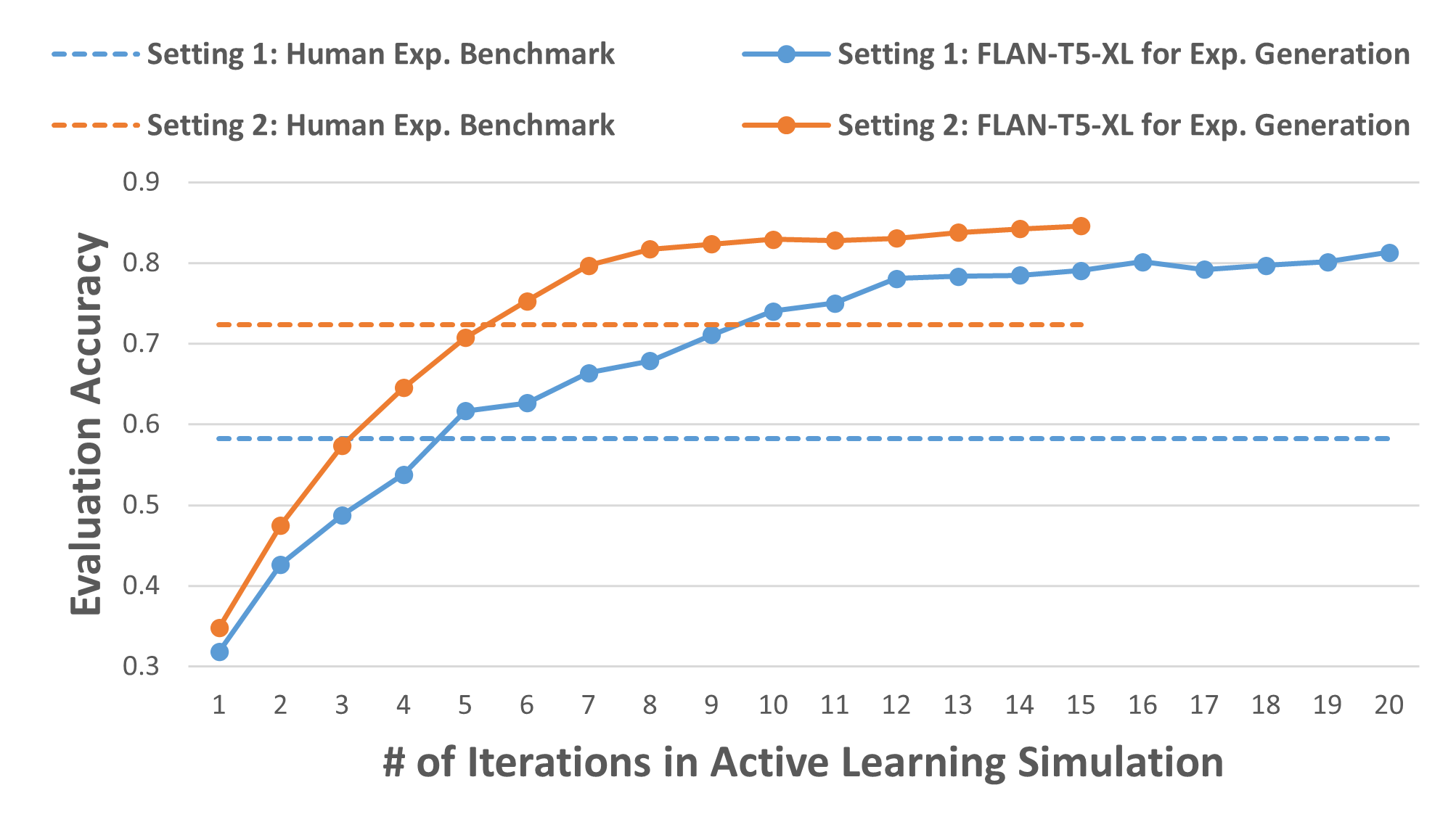}%
    }
    \hfill
    \subfloat[ECQA Dataset]{%
        \includegraphics[clip,width=.95\columnwidth]{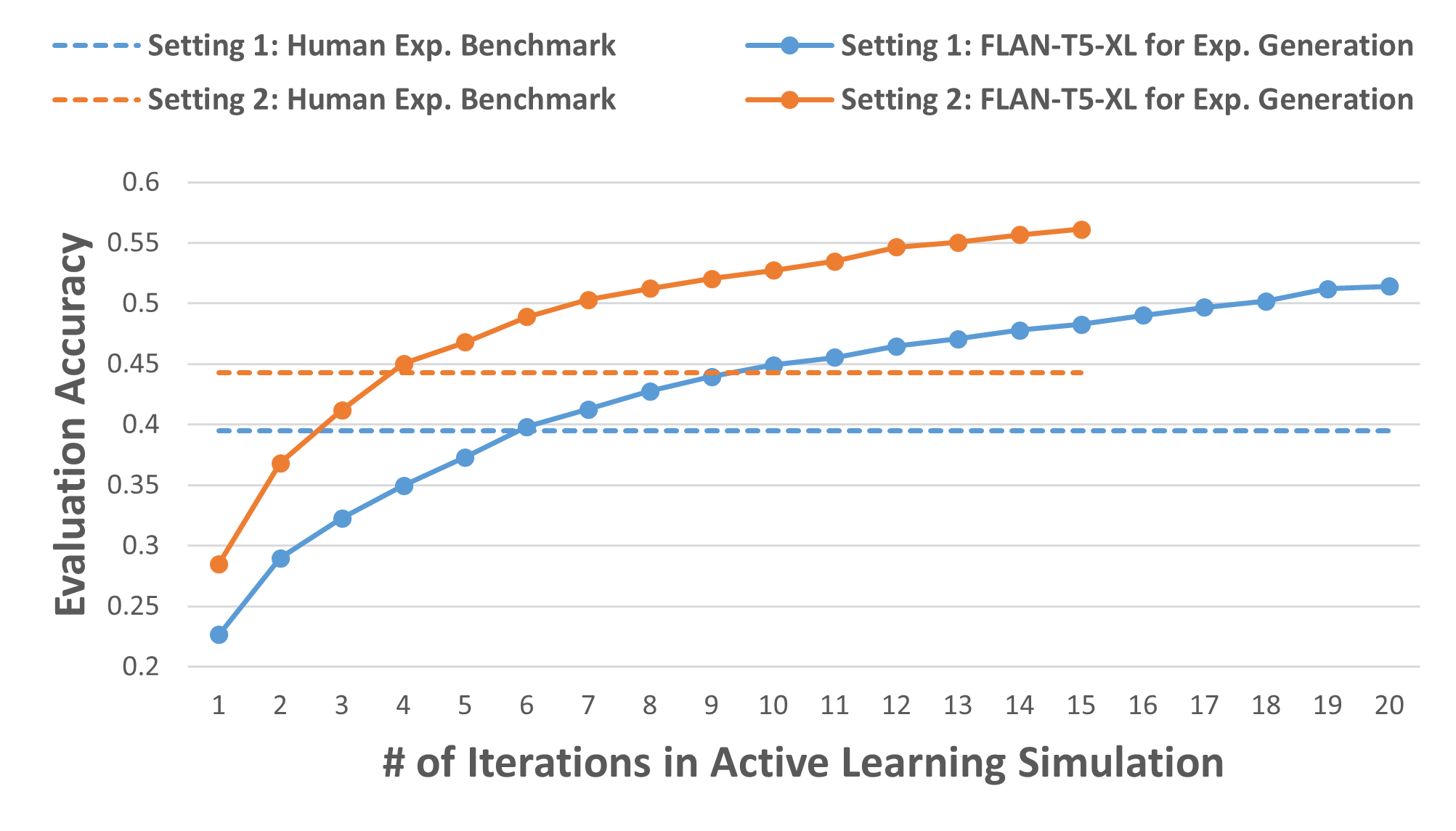}%
    }
    \caption{ Ablation study results of AL simulation experiment with FLAN-T5-XL for explanation (exp.) generation in our Dual-model framework compared with best human-annotated explanations on e-SNLI (top) and ECQA (bottom) datasets. }
    
    \vspace{-1em}
    \label{fig:al_result_llm}
\end{figure}

The recent prevalence of instructional-finetuned large language models (LLMs)~\cite{wei2021finetuned, chowdhery2022palm, ouyang2022training} with exceptional generation capabilities off-the-shelf enabled a straightforward idea upon our dual-model framework: \textbf{can LLMs generate natural language explanations that are on par or even of higher quality than human-annotated ones}, to facilitate the prediction model fine-tuning process? 
We conduct ablation experiments to leverage FLAN-T5-XL~\cite{chung2022scaling} for explanation generation in our framework to substitute the T5 model fine-tuned on human explanations (LLM-AL, hereinafter).
We conduct the AL simulations on e-SNLI and ECQA datasets to explore whether we can further reduce human annotation efforts.

The results are presented in Figure~\ref{fig:al_result_llm}, where a horizontal dotted line represents the benchmark of the explanation generation model fine-tuned on human-annotated explanations in Section~\ref{sec:als_result} and~\ref{sec:abla_ecqa}. 
The LLM-AL framework significantly outperforms the explanation generation model guided by human annotation in both Active Learning settings. 
However, \textbf{we hypothesize the LLM’s explanation generation capability can vary from task to task}.
It may be highly efficient in relatively easy tasks, such as e-SNLI and ECQA datasets, both of which are training datasets for FLAN-T5. Yet, LLMs may struggle to provide helpful explanations in complex real-world domain-specific tasks, where human experts' feedback may still be necessary and preferred. This leads to another potential avenue for future work: exploring the capability and limitations of leveraging LLMs for explanation generation in real-world scenarios.
\section{Conclusion and Future Work}










In summary, this paper introduces a novel dual-model AL system designed to address the common real-world need for domain experts to provide both classification labels and natural language explanations.
Our system comprises a purpose-built data diversity-based AL example selector and two sequence-to-sequence language models, one for explanation generation and the other for label prediction.
Through an AL simulation evaluation and a human assessment of the e-SNLI dataset, our results demonstrate the effectiveness of explanations in AL sampling with our system. They consistently outperform both baselines, and the explanations generated by our system are preferred over a state-of-the-art explanation-generation system.

Our work lays a step-stone towards a human-centered interactive AI solution (it can be easily implemented as an interactive system as illustrated in Fig~\ref{fig:system_arch_future} in Appendix~\ref{app:system_arch_future}) that supports domain experts for their data annotation tasks.
Many real-world tasks still require domain experts to review and annotate each data instance with a decision and an explanation for accountability purposes (e.g., a lawyer reviewing and signing off on a legal document). We invite fellow researchers to join us in advancing this research direction, essential for supporting this prevalent real-world requirement.

\section{Limitations}
In this paper, we demonstrate the effectiveness of our framework on a representative large-scale classification dataset (e-SNLI), but there are many other NLP tasks, such as question answering and commonsense reasoning.
The generalizability of our system on other NLP tasks remains unexplored.
Another limitation is that this work proposed a data diversity-based AL selector design. We benchmark it with a traditional data diversity-based selector as well as a model probability-based design to demonstrate the usefulness of explanations.
Prior literature has proposed other designs, such as ensemble approaches, which are not evaluated in this paper. 



\section*{Acknowledgements}

This work was supported by the Rensselaer-IBM
AI Research Collaboration (http://airc.rpi.edu), which is part of the IBM AI Horizons Network (http://ibm.biz/AIHorizons).

\bibliography{anthology, custom}

\begin{thebibliography}{70}
\expandafter\ifx\csname natexlab\endcsname\relax\def\natexlab#1{#1}\fi

\bibitem[{Aggarwal et~al.(2021)Aggarwal, Mandowara, Agrawal, Khandelwal,
  Singla, and Garg}]{aggarwal-etal-2021-explanations}
Shourya Aggarwal, Divyanshu Mandowara, Vishwajeet Agrawal, Dinesh Khandelwal,
  Parag Singla, and Dinesh Garg. 2021.
\newblock \href {https://doi.org/10.18653/v1/2021.acl-long.238} {{E}xplanations
  for {C}ommonsense{QA}: {N}ew {D}ataset and {M}odels}.
\newblock In \emph{Proceedings of the 59th Annual Meeting of the Association
  for Computational Linguistics and the 11th International Joint Conference on
  Natural Language Processing (Volume 1: Long Papers)}, pages 3050--3065,
  Online. Association for Computational Linguistics.

\bibitem[{Ash et~al.(2019)Ash, Zhang, Krishnamurthy, Langford, and
  Agarwal}]{ash2019deep}
Jordan~T Ash, Chicheng Zhang, Akshay Krishnamurthy, John Langford, and Alekh
  Agarwal. 2019.
\newblock Deep batch active learning by diverse, uncertain gradient lower
  bounds.
\newblock \emph{arXiv preprint arXiv:1906.03671}.

\bibitem[{Bhat et~al.(2021)Bhat, Sordoni, and Mukherjee}]{bhat-etal-2021-self}
Meghana~Moorthy Bhat, Alessandro Sordoni, and Subhabrata Mukherjee. 2021.
\newblock \href {https://doi.org/10.18653/v1/2021.emnlp-main.836}
  {Self-training with few-shot rationalization}.
\newblock In \emph{Proceedings of the 2021 Conference on Empirical Methods in
  Natural Language Processing}, pages 10702--10712, Online and Punta Cana,
  Dominican Republic. Association for Computational Linguistics.

\bibitem[{Bowman et~al.(2015)Bowman, Angeli, Potts, and
  Manning}]{bowman-etal-2015-large}
Samuel~R. Bowman, Gabor Angeli, Christopher Potts, and Christopher~D. Manning.
  2015.
\newblock \href {https://doi.org/10.18653/v1/D15-1075} {A large annotated
  corpus for learning natural language inference}.
\newblock In \emph{Proceedings of the 2015 Conference on Empirical Methods in
  Natural Language Processing}, pages 632--642, Lisbon, Portugal. Association
  for Computational Linguistics.

\bibitem[{Bragg et~al.(2021)Bragg, Cohan, Lo, and Beltagy}]{bragg2021flex}
Jonathan Bragg, Arman Cohan, Kyle Lo, and Iz~Beltagy. 2021.
\newblock Flex: Unifying evaluation for few-shot nlp.
\newblock \emph{Advances in Neural Information Processing Systems},
  34:15787--15800.

\bibitem[{Camburu et~al.(2018)Camburu, Rockt{\"a}schel, Lukasiewicz, and
  Blunsom}]{camburu2018snli}
Oana-Maria Camburu, Tim Rockt{\"a}schel, Thomas Lukasiewicz, and Phil Blunsom.
  2018.
\newblock \href
  {https://papers.nips.cc/paper_files/paper/2018/hash/4c7a167bb329bd92580a99ce422d6fa6-Abstract.html}
  {{e-SNLI:} {N}atural language inference with natural language explanations}.
\newblock \emph{Advances in Neural Information Processing Systems}, 31.

\bibitem[{Carton et~al.(2020)Carton, Rathore, and
  Tan}]{carton-etal-2020-evaluating}
Samuel Carton, Anirudh Rathore, and Chenhao Tan. 2020.
\newblock \href {https://doi.org/10.18653/v1/2020.emnlp-main.747} {Evaluating
  and characterizing human rationales}.
\newblock In \emph{Proceedings of the 2020 Conference on Empirical Methods in
  Natural Language Processing (EMNLP)}, pages 9294--9307, Online. Association
  for Computational Linguistics.

\bibitem[{Chen et~al.(2021)Chen, Chen, Shi, and Zhang}]{chen2021generate}
Hanxiong Chen, Xu~Chen, Shaoyun Shi, and Yongfeng Zhang. 2021.
\newblock Generate natural language explanations for recommendation.
\newblock \emph{arXiv preprint arXiv:2101.03392}.

\bibitem[{Chen et~al.(2022)Chen, He, Narasimhan, and
  Chen}]{chen-etal-2022-rationalization}
Howard Chen, Jacqueline He, Karthik Narasimhan, and Danqi Chen. 2022.
\newblock \href {https://doi.org/10.18653/v1/2022.naacl-main.278} {Can
  rationalization improve robustness?}
\newblock In \emph{Proceedings of the 2022 Conference of the North American
  Chapter of the Association for Computational Linguistics: Human Language
  Technologies}, pages 3792--3805, Seattle, United States. Association for
  Computational Linguistics.

\bibitem[{Chmielewski and Kucker(2020)}]{chmielewski2020mturk}
Michael Chmielewski and Sarah~C Kucker. 2020.
\newblock An mturk crisis? shifts in data quality and the impact on study
  results.
\newblock \emph{Social Psychological and Personality Science}, 11(4):464--473.

\bibitem[{Chowdhery et~al.(2022)Chowdhery, Narang, Devlin, Bosma, Mishra,
  Roberts, Barham, Chung, Sutton, Gehrmann et~al.}]{chowdhery2022palm}
Aakanksha Chowdhery, Sharan Narang, Jacob Devlin, Maarten Bosma, Gaurav Mishra,
  Adam Roberts, Paul Barham, Hyung~Won Chung, Charles Sutton, Sebastian
  Gehrmann, et~al. 2022.
\newblock Palm: Scaling language modeling with pathways.
\newblock \emph{arXiv preprint arXiv:2204.02311}.

\bibitem[{Chung et~al.(2022)Chung, Hou, Longpre, Zoph, Tay, Fedus, Li, Wang,
  Dehghani, Brahma et~al.}]{chung2022scaling}
Hyung~Won Chung, Le~Hou, Shayne Longpre, Barret Zoph, Yi~Tay, William Fedus,
  Eric Li, Xuezhi Wang, Mostafa Dehghani, Siddhartha Brahma, et~al. 2022.
\newblock Scaling instruction-finetuned language models.
\newblock \emph{arXiv preprint arXiv:2210.11416}.

\bibitem[{Dalvi et~al.(2022)Dalvi, Tafjord, and Clark}]{dalvi2022towards}
Bhavana Dalvi, Oyvind Tafjord, and Peter Clark. 2022.
\newblock Towards teachable reasoning systems.
\newblock \emph{arXiv preprint arXiv:2204.13074}.

\bibitem[{Devlin et~al.(2019)Devlin, Chang, Lee, and
  Toutanova}]{devlin-etal-2019-bert}
Jacob Devlin, Ming-Wei Chang, Kenton Lee, and Kristina Toutanova. 2019.
\newblock \href {https://doi.org/10.18653/v1/N19-1423} {{BERT}: Pre-training of
  deep bidirectional transformers for language understanding}.
\newblock In \emph{Proceedings of the 2019 Conference of the North {A}merican
  Chapter of the Association for Computational Linguistics: Human Language
  Technologies, Volume 1 (Long and Short Papers)}, pages 4171--4186,
  Minneapolis, Minnesota. Association for Computational Linguistics.

\bibitem[{Drozdal et~al.(2020)Drozdal, Weisz, Wang, Dass, Yao, Zhao, Muller,
  Ju, and Su}]{drozdal2020trust}
Jaimie Drozdal, Justin Weisz, Dakuo Wang, Gaurav Dass, Bingsheng Yao, Changruo
  Zhao, Michael Muller, Lin Ju, and Hui Su. 2020.
\newblock Trust in automl: exploring information needs for establishing trust
  in automated machine learning systems.
\newblock In \emph{Proceedings of the 25th international conference on
  intelligent user interfaces}, pages 297--307.

\bibitem[{Duan et~al.(2017)Duan, Tang, Chen, and
  Zhou}]{duan-etal-2017-question}
Nan Duan, Duyu Tang, Peng Chen, and Ming Zhou. 2017.
\newblock \href {https://doi.org/10.18653/v1/D17-1090} {Question generation for
  question answering}.
\newblock In \emph{Proceedings of the 2017 Conference on Empirical Methods in
  Natural Language Processing}, pages 866--874, Copenhagen, Denmark.
  Association for Computational Linguistics.

\bibitem[{Elliott and Woodward(2007)}]{elliott2007statistical}
Alan~C Elliott and Wayne~A Woodward. 2007.
\newblock \emph{Statistical analysis quick reference guidebook: With SPSS
  examples}.
\newblock Sage.

\bibitem[{Fu et~al.(2013)Fu, Zhu, and Li}]{fu2013survey}
Yifan Fu, Xingquan Zhu, and Bin Li. 2013.
\newblock A survey on instance selection for active learning.
\newblock \emph{Knowledge and information systems}, 35:249--283.

\bibitem[{Gao et~al.(2021)Gao, Fisch, and Chen}]{gao-etal-2021-making}
Tianyu Gao, Adam Fisch, and Danqi Chen. 2021.
\newblock \href {https://doi.org/10.18653/v1/2021.acl-long.295} {Making
  pre-trained language models better few-shot learners}.
\newblock In \emph{Proceedings of the 59th Annual Meeting of the Association
  for Computational Linguistics and the 11th International Joint Conference on
  Natural Language Processing (Volume 1: Long Papers)}, pages 3816--3830,
  Online. Association for Computational Linguistics.

\bibitem[{Geva et~al.(2019)Geva, Goldberg, and Berant}]{geva2019we}
Mor Geva, Yoav Goldberg, and Jonathan Berant. 2019.
\newblock Are we modeling the task or the annotator? an investigation of
  annotator bias in natural language understanding datasets.
\newblock \emph{arXiv preprint arXiv:1908.07898}.

\bibitem[{Hase and Bansal(2022)}]{hase-bansal-2022-models}
Peter Hase and Mohit Bansal. 2022.
\newblock \href {https://doi.org/10.18653/v1/2022.lnls-1.4} {When can models
  learn from explanations? a formal framework for understanding the roles of
  explanation data}.
\newblock In \emph{Proceedings of the First Workshop on Learning with Natural
  Language Supervision}, pages 29--39, Dublin, Ireland. Association for
  Computational Linguistics.

\bibitem[{Kasai et~al.(2019)Kasai, Qian, Gurajada, Li, and
  Popa}]{kasai-etal-2019-low}
Jungo Kasai, Kun Qian, Sairam Gurajada, Yunyao Li, and Lucian Popa. 2019.
\newblock \href {https://doi.org/10.18653/v1/P19-1586} {Low-resource deep
  entity resolution with transfer and active learning}.
\newblock In \emph{Proceedings of the 57th Annual Meeting of the Association
  for Computational Linguistics}, pages 5851--5861, Florence, Italy.
  Association for Computational Linguistics.

\bibitem[{Ko{\v{c}}isk{\'y} et~al.(2018)Ko{\v{c}}isk{\'y}, Schwarz, Blunsom,
  Dyer, Hermann, Melis, and Grefenstette}]{kocisky-etal-2018-narrativeqa}
Tom{\'a}{\v{s}} Ko{\v{c}}isk{\'y}, Jonathan Schwarz, Phil Blunsom, Chris Dyer,
  Karl~Moritz Hermann, G{\'a}bor Melis, and Edward Grefenstette. 2018.
\newblock \href {https://doi.org/10.1162/tacl_a_00023} {The {N}arrative{QA}
  reading comprehension challenge}.
\newblock \emph{Transactions of the Association for Computational Linguistics},
  6:317--328.

\bibitem[{Kumar and Talukdar(2020)}]{kumar-talukdar-2020-nile}
Sawan Kumar and Partha Talukdar. 2020.
\newblock \href {https://doi.org/10.18653/v1/2020.acl-main.771} {{NILE} :
  Natural language inference with faithful natural language explanations}.
\newblock In \emph{Proceedings of the 58th Annual Meeting of the Association
  for Computational Linguistics}, pages 8730--8742, Online. Association for
  Computational Linguistics.

\bibitem[{Latcinnik and Berant(2020)}]{latcinnik2020explaining}
Veronica Latcinnik and Jonathan Berant. 2020.
\newblock Explaining question answering models through text generation.
\newblock \emph{arXiv preprint arXiv:2004.05569}.

\bibitem[{Lewis et~al.(2020)Lewis, Liu, Goyal, Ghazvininejad, Mohamed, Levy,
  Stoyanov, and Zettlemoyer}]{lewis-etal-2020-bart}
Mike Lewis, Yinhan Liu, Naman Goyal, Marjan Ghazvininejad, Abdelrahman Mohamed,
  Omer Levy, Veselin Stoyanov, and Luke Zettlemoyer. 2020.
\newblock \href {https://doi.org/10.18653/v1/2020.acl-main.703} {{BART}:
  Denoising sequence-to-sequence pre-training for natural language generation,
  translation, and comprehension}.
\newblock In \emph{Proceedings of the 58th Annual Meeting of the Association
  for Computational Linguistics}, pages 7871--7880, Online. Association for
  Computational Linguistics.

\bibitem[{Lipton(2018)}]{lipton2018mythos}
Zachary~C Lipton. 2018.
\newblock The mythos of model interpretability: In machine learning, the
  concept of interpretability is both important and slippery.
\newblock \emph{Queue}, 16(3):31--57.

\bibitem[{Liu et~al.(2022)Liu, Liu, Lu, Welleck, West, Le~Bras, Choi, and
  Hajishirzi}]{liu-etal-2022-generated}
Jiacheng Liu, Alisa Liu, Ximing Lu, Sean Welleck, Peter West, Ronan Le~Bras,
  Yejin Choi, and Hannaneh Hajishirzi. 2022.
\newblock \href {https://doi.org/10.18653/v1/2022.acl-long.225} {Generated
  knowledge prompting for commonsense reasoning}.
\newblock In \emph{Proceedings of the 60th Annual Meeting of the Association
  for Computational Linguistics (Volume 1: Long Papers)}, pages 3154--3169,
  Dublin, Ireland. Association for Computational Linguistics.

\bibitem[{Lundberg and Lee(2017)}]{lundberg2017unified}
Scott~M Lundberg and Su-In Lee. 2017.
\newblock A unified approach to interpreting model predictions.
\newblock \emph{Advances in neural information processing systems}, 30.

\bibitem[{Marasovic et~al.(2022)Marasovic, Beltagy, Downey, and
  Peters}]{marasovic-etal-2022-shot}
Ana Marasovic, Iz~Beltagy, Doug Downey, and Matthew Peters. 2022.
\newblock \href {https://doi.org/10.18653/v1/2022.findings-naacl.31} {Few-shot
  self-rationalization with natural language prompts}.
\newblock In \emph{Findings of the Association for Computational Linguistics:
  NAACL 2022}, pages 410--424, Seattle, United States. Association for
  Computational Linguistics.

\bibitem[{Narang et~al.(2020)Narang, Raffel, Lee, Roberts, Fiedel, and
  Malkan}]{narang2020wt5}
Sharan Narang, Colin Raffel, Katherine Lee, Adam Roberts, Noah Fiedel, and
  Karishma Malkan. 2020.
\newblock Wt5?! training text-to-text models to explain their predictions.
\newblock \emph{arXiv preprint arXiv:2004.14546}.

\bibitem[{Nguyen and Smeulders(2004)}]{nguyen2004active}
Hieu~T Nguyen and Arnold Smeulders. 2004.
\newblock Active learning using pre-clustering.
\newblock In \emph{Proceedings of the twenty-first international conference on
  Machine learning}, page~79.

\bibitem[{Olsson(2009)}]{olsson2009literature}
Fredrik Olsson. 2009.
\newblock A literature survey of active machine learning in the context of
  natural language processing.

\bibitem[{Ouyang et~al.(2022)Ouyang, Wu, Jiang, Almeida, Wainwright, Mishkin,
  Zhang, Agarwal, Slama, Ray et~al.}]{ouyang2022training}
Long Ouyang, Jeffrey Wu, Xu~Jiang, Diogo Almeida, Carroll Wainwright, Pamela
  Mishkin, Chong Zhang, Sandhini Agarwal, Katarina Slama, Alex Ray, et~al.
  2022.
\newblock Training language models to follow instructions with human feedback.
\newblock \emph{Advances in Neural Information Processing Systems},
  35:27730--27744.

\bibitem[{Paranjape et~al.(2021)Paranjape, Michael, Ghazvininejad, Hajishirzi,
  and Zettlemoyer}]{paranjape-etal-2021-prompting}
Bhargavi Paranjape, Julian Michael, Marjan Ghazvininejad, Hannaneh Hajishirzi,
  and Luke Zettlemoyer. 2021.
\newblock \href {https://doi.org/10.18653/v1/2021.findings-acl.366} {Prompting
  contrastive explanations for commonsense reasoning tasks}.
\newblock In \emph{Findings of the Association for Computational Linguistics:
  ACL-IJCNLP 2021}, pages 4179--4192, Online. Association for Computational
  Linguistics.

\bibitem[{Radford et~al.(2019)Radford, Wu, Child, Luan, Amodei, Sutskever
  et~al.}]{radford2019language}
Alec Radford, Jeffrey Wu, Rewon Child, David Luan, Dario Amodei, Ilya
  Sutskever, et~al. 2019.
\newblock Language models are unsupervised multitask learners.
\newblock \emph{OpenAI blog}, 1(8):9.

\bibitem[{Raffel et~al.(2020)Raffel, Shazeer, Roberts, Lee, Narang, Matena,
  Zhou, Li, Liu et~al.}]{raffel2020exploring}
Colin Raffel, Noam Shazeer, Adam Roberts, Katherine Lee, Sharan Narang, Michael
  Matena, Yanqi Zhou, Wei Li, Peter~J Liu, et~al. 2020.
\newblock Exploring the limits of transfer learning with a unified text-to-text
  transformer.
\newblock \emph{J. Mach. Learn. Res.}, 21(140):1--67.

\bibitem[{Rajagopal et~al.(2021)Rajagopal, Balachandran, Hovy, and
  Tsvetkov}]{rajagopal-etal-2021-selfexplain}
Dheeraj Rajagopal, Vidhisha Balachandran, Eduard~H Hovy, and Yulia Tsvetkov.
  2021.
\newblock \href {https://doi.org/10.18653/v1/2021.emnlp-main.64}
  {{SELFEXPLAIN}: A self-explaining architecture for neural text classifiers}.
\newblock In \emph{Proceedings of the 2021 Conference on Empirical Methods in
  Natural Language Processing}, pages 836--850, Online and Punta Cana,
  Dominican Republic. Association for Computational Linguistics.

\bibitem[{Rajani et~al.(2019)Rajani, McCann, Xiong, and
  Socher}]{rajani-etal-2019-explain}
Nazneen~Fatema Rajani, Bryan McCann, Caiming Xiong, and Richard Socher. 2019.
\newblock \href {https://doi.org/10.18653/v1/P19-1487} {Explain yourself!
  leveraging language models for commonsense reasoning}.
\newblock In \emph{Proceedings of the 57th Annual Meeting of the Association
  for Computational Linguistics}, pages 4932--4942, Florence, Italy.
  Association for Computational Linguistics.

\bibitem[{Rajpurkar et~al.(2016)Rajpurkar, Zhang, Lopyrev, and
  Liang}]{rajpurkar-etal-2016-squad}
Pranav Rajpurkar, Jian Zhang, Konstantin Lopyrev, and Percy Liang. 2016.
\newblock \href {https://doi.org/10.18653/v1/D16-1264} {{SQ}u{AD}: 100,000+
  questions for machine comprehension of text}.
\newblock In \emph{Proceedings of the 2016 Conference on Empirical Methods in
  Natural Language Processing}, pages 2383--2392, Austin, Texas. Association
  for Computational Linguistics.

\bibitem[{Reimers and Gurevych(2019)}]{reimers-2019-sentence-bert}
Nils Reimers and Iryna Gurevych. 2019.
\newblock \href {https://arxiv.org/abs/1908.10084} {Sentence-bert: Sentence
  embeddings using siamese bert-networks}.
\newblock In \emph{Proceedings of the 2019 Conference on Empirical Methods in
  Natural Language Processing}. Association for Computational Linguistics.

\bibitem[{Ren et~al.(2021)Ren, Xiao, Chang, Huang, Li, Gupta, Chen, and
  Wang}]{ren2021survey}
Pengzhen Ren, Yun Xiao, Xiaojun Chang, Po-Yao Huang, Zhihui Li, Brij~B Gupta,
  Xiaojiang Chen, and Xin Wang. 2021.
\newblock A survey of deep active learning.
\newblock \emph{ACM computing surveys (CSUR)}, 54(9):1--40.

\bibitem[{Ribeiro et~al.(2016)Ribeiro, Singh, and Guestrin}]{ribeiro2016should}
Marco~Tulio Ribeiro, Sameer Singh, and Carlos Guestrin. 2016.
\newblock " why should i trust you?" explaining the predictions of any
  classifier.
\newblock In \emph{Proceedings of the 22nd ACM SIGKDD international conference
  on knowledge discovery and data mining}, pages 1135--1144.

\bibitem[{Schick and Sch{\"u}tze(2021)}]{schick-schutze-2021-exploiting}
Timo Schick and Hinrich Sch{\"u}tze. 2021.
\newblock \href {https://aclanthology.org/2021.eacl-main.20} {Exploiting
  cloze-questions for few-shot text classification and natural language
  inference}.
\newblock In \emph{Proceedings of the 16th Conference of the European Chapter
  of the Association for Computational Linguistics: Main Volume}, pages
  255--269, Online. Association for Computational Linguistics.

\bibitem[{Schr{\"o}der and Niekler(2020)}]{schroder2020survey}
Christopher Schr{\"o}der and Andreas Niekler. 2020.
\newblock A survey of active learning for text classification using deep neural
  networks.
\newblock \emph{arXiv preprint arXiv:2008.07267}.

\bibitem[{Sener and Savarese(2017)}]{sener2017active}
Ozan Sener and Silvio Savarese. 2017.
\newblock Active learning for convolutional neural networks: A core-set
  approach.
\newblock \emph{arXiv preprint arXiv:1708.00489}.

\bibitem[{Settles(2009)}]{settles2009active}
Burr Settles. 2009.
\newblock Active learning literature survey.

\bibitem[{Sharma et~al.(2015)Sharma, Zhuang, and
  Bilgic}]{sharma-etal-2015-active}
Manali Sharma, Di~Zhuang, and Mustafa Bilgic. 2015.
\newblock \href {https://doi.org/10.3115/v1/N15-1047} {Active learning with
  rationales for text classification}.
\newblock In \emph{Proceedings of the 2015 Conference of the North {A}merican
  Chapter of the Association for Computational Linguistics: Human Language
  Technologies}, pages 441--451, Denver, Colorado. Association for
  Computational Linguistics.

\bibitem[{Shen et~al.(2017)Shen, Yun, Lipton, Kronrod, and
  Anandkumar}]{shen-etal-2017-deep}
Yanyao Shen, Hyokun Yun, Zachary Lipton, Yakov Kronrod, and Animashree
  Anandkumar. 2017.
\newblock \href {https://doi.org/10.18653/v1/W17-2630} {Deep active learning
  for named entity recognition}.
\newblock In \emph{Proceedings of the 2nd Workshop on Representation Learning
  for {NLP}}, pages 252--256, Vancouver, Canada. Association for Computational
  Linguistics.

\bibitem[{Sun et~al.(2022)Sun, Swayamdipta, May, and Ma}]{sun2022investigating}
Jiao Sun, Swabha Swayamdipta, Jonathan May, and Xuezhe Ma. 2022.
\newblock Investigating the benefits of free-form rationales.
\newblock \emph{arXiv preprint arXiv:2206.11083}.

\bibitem[{Tafjord et~al.(2021)Tafjord, Dalvi, and
  Clark}]{tafjord-etal-2021-proofwriter}
Oyvind Tafjord, Bhavana Dalvi, and Peter Clark. 2021.
\newblock \href {https://doi.org/10.18653/v1/2021.findings-acl.317}
  {{P}roof{W}riter: Generating implications, proofs, and abductive statements
  over natural language}.
\newblock In \emph{Findings of the Association for Computational Linguistics:
  ACL-IJCNLP 2021}, pages 3621--3634, Online. Association for Computational
  Linguistics.

\bibitem[{Talmor et~al.(2019)Talmor, Herzig, Lourie, and
  Berant}]{talmor-etal-2019-commonsenseqa}
Alon Talmor, Jonathan Herzig, Nicholas Lourie, and Jonathan Berant. 2019.
\newblock \href {https://doi.org/10.18653/v1/N19-1421} {{C}ommonsense{QA}: A
  question answering challenge targeting commonsense knowledge}.
\newblock In \emph{Proceedings of the 2019 Conference of the North {A}merican
  Chapter of the Association for Computational Linguistics: Human Language
  Technologies, Volume 1 (Long and Short Papers)}, pages 4149--4158,
  Minneapolis, Minnesota. Association for Computational Linguistics.

\bibitem[{Talmor et~al.(2020)Talmor, Tafjord, Clark, Goldberg, and
  Berant}]{talmor2020leap}
Alon Talmor, Oyvind Tafjord, Peter Clark, Yoav Goldberg, and Jonathan Berant.
  2020.
\newblock Leap-of-thought: Teaching pre-trained models to systematically reason
  over implicit knowledge.
\newblock \emph{Advances in Neural Information Processing Systems},
  33:20227--20237.

\bibitem[{Teso and Kersting(2019)}]{teso2019explanatory}
Stefano Teso and Kristian Kersting. 2019.
\newblock Explanatory interactive machine learning.
\newblock In \emph{Proceedings of the 2019 AAAI/ACM Conference on AI, Ethics,
  and Society}, pages 239--245.

\bibitem[{Wang et~al.(2018)Wang, Singh, Michael, Hill, Levy, and
  Bowman}]{wang-etal-2018-glue}
Alex Wang, Amanpreet Singh, Julian Michael, Felix Hill, Omer Levy, and Samuel
  Bowman. 2018.
\newblock \href {https://doi.org/10.18653/v1/W18-5446} {{GLUE}: A multi-task
  benchmark and analysis platform for natural language understanding}.
\newblock In \emph{Proceedings of the 2018 {EMNLP} Workshop {B}lackbox{NLP}:
  Analyzing and Interpreting Neural Networks for {NLP}}, pages 353--355,
  Brussels, Belgium. Association for Computational Linguistics.

\bibitem[{Wei et~al.(2021)Wei, Bosma, Zhao, Guu, Yu, Lester, Du, Dai, and
  Le}]{wei2021finetuned}
Jason Wei, Maarten Bosma, Vincent~Y Zhao, Kelvin Guu, Adams~Wei Yu, Brian
  Lester, Nan Du, Andrew~M Dai, and Quoc~V Le. 2021.
\newblock Finetuned language models are zero-shot learners.
\newblock \emph{arXiv preprint arXiv:2109.01652}.

\bibitem[{Wiegreffe and Marasovic(2021)}]{wiegreffe2021teach}
Sarah Wiegreffe and Ana Marasovic. 2021.
\newblock Teach me to explain: A review of datasets for explainable natural
  language processing.
\newblock In \emph{Thirty-fifth Conference on Neural Information Processing
  Systems Datasets and Benchmarks Track (Round 1)}.

\bibitem[{Wiegreffe et~al.(2021)Wiegreffe, Marasovi{\'c}, and
  Smith}]{wiegreffe-etal-2021-measuring}
Sarah Wiegreffe, Ana Marasovi{\'c}, and Noah~A. Smith. 2021.
\newblock \href {https://doi.org/10.18653/v1/2021.emnlp-main.804} {{M}easuring
  association between labels and free-text rationales}.
\newblock In \emph{Proceedings of the 2021 Conference on Empirical Methods in
  Natural Language Processing}, pages 10266--10284, Online and Punta Cana,
  Dominican Republic. Association for Computational Linguistics.

\bibitem[{Williams et~al.(2018)Williams, Nangia, and
  Bowman}]{williams-etal-2018-broad}
Adina Williams, Nikita Nangia, and Samuel Bowman. 2018.
\newblock \href {https://doi.org/10.18653/v1/N18-1101} {A broad-coverage
  challenge corpus for sentence understanding through inference}.
\newblock In \emph{Proceedings of the 2018 Conference of the North {A}merican
  Chapter of the Association for Computational Linguistics: Human Language
  Technologies, Volume 1 (Long Papers)}, pages 1112--1122, New Orleans,
  Louisiana. Association for Computational Linguistics.

\bibitem[{Winata et~al.(2021)Winata, Madotto, Lin, Liu, Yosinski, and
  Fung}]{winata-etal-2021-language}
Genta~Indra Winata, Andrea Madotto, Zhaojiang Lin, Rosanne Liu, Jason Yosinski,
  and Pascale Fung. 2021.
\newblock \href {https://doi.org/10.18653/v1/2021.mrl-1.1} {Language models are
  few-shot multilingual learners}.
\newblock In \emph{Proceedings of the 1st Workshop on Multilingual
  Representation Learning}, pages 1--15, Punta Cana, Dominican Republic.
  Association for Computational Linguistics.

\bibitem[{Xu et~al.(2022)Xu, Wang, Yu, Ritchie, Yao, Wu, Zhang, Li, Bradford,
  Sun et~al.}]{xu2022fantastic}
Ying Xu, Dakuo Wang, Mo~Yu, Daniel Ritchie, Bingsheng Yao, Tongshuang Wu, Zheng
  Zhang, Toby Jia-Jun Li, Nora Bradford, Branda Sun, et~al. 2022.
\newblock Fantastic questions and where to find them: Fairytaleqa--an authentic
  dataset for narrative comprehension.
\newblock \emph{ACL’22}.

\bibitem[{Xu et~al.(2003)Xu, Yu, Tresp, Xu, and Wang}]{xu2003representative}
Zhao Xu, Kai Yu, Volker Tresp, Xiaowei Xu, and Jizhi Wang. 2003.
\newblock Representative sampling for text classification using support vector
  machines.
\newblock In \emph{European conference on information retrieval}, pages
  393--407. Springer.

\bibitem[{Yao et~al.(2023)Yao, Sen, Popa, Hendler, and Wang}]{yao2023human}
Bingsheng Yao, Prithviraj Sen, Lucian Popa, James Hendler, and Dakuo Wang.
  2023.
\newblock Are human explanations always helpful? towards objective evaluation
  of human natural language explanations.
\newblock \emph{arXiv preprint arXiv:2305.03117}.

\bibitem[{Yao et~al.(2022)Yao, Wang, Wu, Zhang, Li, Yu, and
  Xu}]{yao-etal-2022-ais}
Bingsheng Yao, Dakuo Wang, Tongshuang Wu, Zheng Zhang, Toby Li, Mo~Yu, and Ying
  Xu. 2022.
\newblock \href {https://doi.org/10.18653/v1/2022.acl-long.54} {It is {AI}{'}s
  turn to ask humans a question: Question-answer pair generation for
  children{'}s story books}.
\newblock In \emph{Proceedings of the 60th Annual Meeting of the Association
  for Computational Linguistics (Volume 1: Long Papers)}, pages 731--744,
  Dublin, Ireland. Association for Computational Linguistics.

\bibitem[{Yu et~al.(2019)Yu, Chang, Zhang, and
  Jaakkola}]{yu-etal-2019-rethinking}
Mo~Yu, Shiyu Chang, Yang Zhang, and Tommi Jaakkola. 2019.
\newblock \href {https://doi.org/10.18653/v1/D19-1420} {Rethinking cooperative
  rationalization: Introspective extraction and complement control}.
\newblock In \emph{Proceedings of the 2019 Conference on Empirical Methods in
  Natural Language Processing and the 9th International Joint Conference on
  Natural Language Processing (EMNLP-IJCNLP)}, pages 4094--4103, Hong Kong,
  China. Association for Computational Linguistics.

\bibitem[{Zelikman et~al.(2022)Zelikman, Mu, Goodman, and
  Wu}]{zelikman2022star}
Eric Zelikman, Jesse Mu, Noah~D Goodman, and Yuhuai~Tony Wu. 2022.
\newblock Star: Self-taught reasoner bootstrapping reasoning with reasoning.

\bibitem[{Zhang et~al.(2023)Zhang, Yu, Xu, Yin, Lu, Yao, Tory, Padilla,
  Caterino, Zhang et~al.}]{zhang2023rethinking}
Shao Zhang, Jianing Yu, Xuhai Xu, Changchang Yin, Yuxuan Lu, Bingsheng Yao,
  Melanie Tory, Lace~M Padilla, Jeffrey Caterino, Ping Zhang, et~al. 2023.
\newblock Rethinking human-ai collaboration in complex medical decision making:
  A case study in sepsis diagnosis.
\newblock \emph{arXiv preprint arXiv:2309.12368}.

\bibitem[{Zhang et~al.(2022)Zhang, Gong, Liu, He, Chen, and
  Zhou}]{zhang-etal-2022-allsh}
Shujian Zhang, Chengyue Gong, Xingchao Liu, Pengcheng He, Weizhu Chen, and
  Mingyuan Zhou. 2022.
\newblock \href {https://doi.org/10.18653/v1/2022.findings-naacl.99} {{ALLSH}:
  Active learning guided by local sensitivity and hardness}.
\newblock In \emph{Findings of the Association for Computational Linguistics:
  NAACL 2022}, pages 1328--1342, Seattle, United States. Association for
  Computational Linguistics.

\bibitem[{Zhang et~al.(2021)Zhang, Genc, Wang, Ahsen, and
  Fan}]{zhang2021effect}
Zhan Zhang, Yegin Genc, Dakuo Wang, Mehmet~Eren Ahsen, and Xiangmin Fan. 2021.
\newblock Effect of ai explanations on human perceptions of patient-facing
  ai-powered healthcare systems.
\newblock \emph{Journal of Medical Systems}, 45(6):64.

\bibitem[{Zhou et~al.(2023)Zhou, Zhang, and Tan}]{zhou-etal-2023-flame}
Yangqiaoyu Zhou, Yiming Zhang, and Chenhao Tan. 2023.
\newblock \href {https://doi.org/10.18653/v1/2023.acl-long.372} {{FL}am{E}:
  Few-shot learning from natural language explanations}.
\newblock In \emph{Proceedings of the 61st Annual Meeting of the Association
  for Computational Linguistics (Volume 1: Long Papers)}, pages 6743--6763,
  Toronto, Canada. Association for Computational Linguistics.

\end{thebibliography}
\bibliographystyle{acl_natbib}

\clearpage
\appendix
\section*{Appendix}

\section{e-SNLI Examples}
\label{sec:esnli}

Table~\ref{tab:esnli} illustrates an example data of each category in the e-SNLI dataset. Every data instance contains a premise and hypothesis along with a human annotated label and free-form explanation.

\begin{table}[!h]
\centering
\small
\begin{tabular}{p{0.45\textwidth}}
    \toprule
    \textcolor{cpre}{\textbf{Premise: }} \textit{This church choir sings to the masses as they sing joyous songs from the book at a church.} \\
    \textcolor{cpre}{\textbf{Hypothesis: }} \textit{The church is filled with song.} \\
    \textcolor{clabel}{\textbf{Label: }} \textit{entailment} \\
    \textcolor{cexp}{\textbf{Human-annotated explanation: }} \textit{``Filled with song'' is a rephrasing of the "choir sings to the masses.} \\
    \midrule
    \midrule
    \textcolor{cpre}{\textbf{Premise: }} \textit{A man playing an electric guitar on stage.} \\
    \textcolor{cpre}{\textbf{Hypothesis: }} \textit{A man is performing for cash.} \\
    \textcolor{clabel}{\textbf{Label: }} \textit{neutral} \\
    \textcolor{cexp}{\textbf{Human-annotated explanation: }} \textit{It is unknown if the man is performing for cash.} \\
    \midrule
    \midrule
    \textcolor{cpre}{\textbf{Premise: }} \textit{A couple walk hand in hand down a street.} \\
    \textcolor{cpre}{\textbf{Hypothesis: }} \textit{A couple is sitting on a bench.} \\
    \textcolor{clabel}{\textbf{Label: }} \textit{contradiction} \\
    \textcolor{cexp}{\textbf{Human-annotated explanation: }} \textit{The couple cannot be walking and sitting a the same time.} \\
    \bottomrule
    
\end{tabular}

\caption{ Sample data of each category in e-SNLI~\citep{camburu2018snli} dataset. }
\label{tab:esnli}
\vspace{-2em}
\end{table}


\section{Transfer Learning Ablation Study Diagrams}
\label{app:tr_results}

\begin{figure}[ht]
    \centering
    \subfloat[Active Learning on Mulit-NLI using explanation-generation model from e-SNLI with Setting 1]{%
        \includegraphics[clip,width=.95\columnwidth]{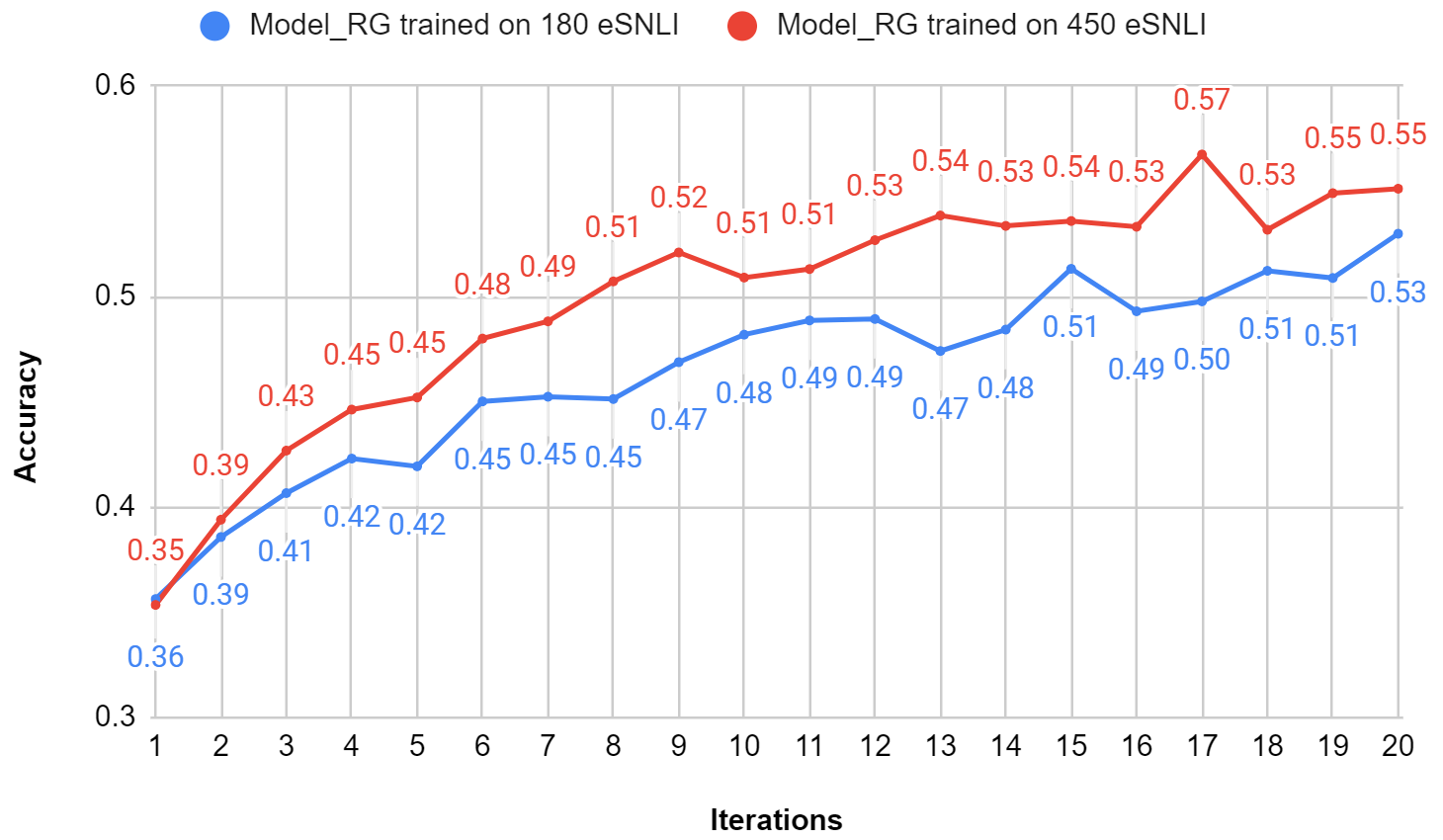}%
    }
    
    \subfloat[Active Learning on Mulit-NLI using explanation-generation model from e-SNLI with Setting 2]{%
        \includegraphics[clip,width=.95\columnwidth]{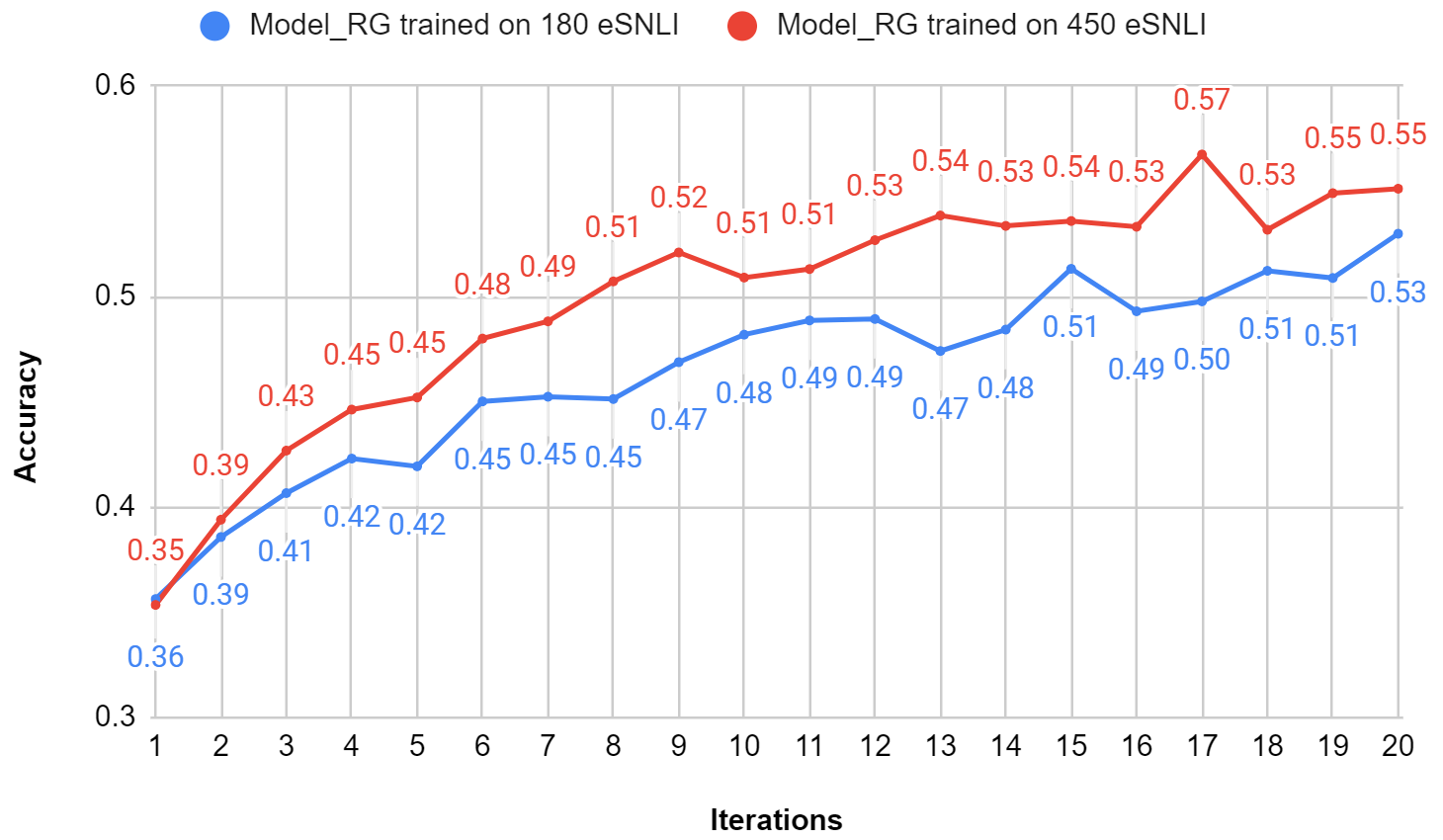}%
    }
    
    \caption{ Results of Transfer Learning Ablation Study of AL Simulation experiment on our Dual-model system from e-SNLI to Multi-NLI. Setting 1/2 refers to the settings for Active Learning Simulation in Section~\ref{sec:als}. }
    
    \vspace{-1em}
    \label{fig:al_tr_result}
\end{figure}

Figure~\ref{fig:al_tr_result} shows the results of our Ablation Study results described in Section~\ref{sec:abla_eval}. The explanation-generation model is fine-tuned from AL on e-SNLI dataset with two different AL settings, then we freeze the explanation-generation model to train the prediction model in AL simulation for Multi-NLI dataset under two settings. Setting 1/2 refers to the settings for Active Learning Simulation in Section~\ref{sec:als}. 

\section{System Environment and Hyper-Parameters}
\label{app:hyperparameters}
The computing resource of all the experiments we conducted in this paper has $128$ Gigabytes of RAM. In addition, we use $2$ NVIDIA Tesla V100 GPU for the preliminary experiment and $8$ NVIDIA Tesla V100 GPU for the AL simulation experiment.

\subsection{Preliminary Experiment}
For the Preliminary experiment described in Section~\ref{sec:pe}, we leverage the same set of fine-tuning hyper-parameters other than the number of fine-tuning epochs for the explanation-generation model (denotes as $M_{EG}$) and the prediction model (denotes as $M_{P}$). The same set of hyper-parameters is: $batch\_size\_per\_GPU=2; learning\_rate=1e^{-4};input\_max\_length=512; target\_max\_length=64$

We conduct a hyper-parameter search for the number of fine-tuning epochs for each amount of sampled examples, details are shown in Table~\ref{tab:finetune_epochs}.

\begin{table}[h]
\centering
\resizebox{.95\columnwidth}{!}{%
\begin{tabular}{lcc}
\toprule
\begin{tabular}[x]{@{}l@{}}\textbf{\# of train data}\\\textbf{per category / total}\end{tabular} & \textbf{epoch for $M_{RG}$} & \textbf{epoch for $M_{P}$} \\
\cmidrule(lr){1-1} \cmidrule(lr){2-2} \cmidrule(lr){3-3}
10 / 30       &   25    &   100    \\
50 / 150      &   25    &   250    \\
100 / 300      &   10    &   250    \\
500 / 1500     &   5    &    50   \\
1500 / 4500     &   5    &    50   \\
3000 / 9000     &   5    &    25   \\
5000 / 15000    &   5    &    25   \\
Full     &   1    &    1   \\

\bottomrule
\end{tabular}
}
\caption{ Fine-tuning epochs of each model in our dual-model system with different data amount settings. }
\label{tab:finetune_epochs}
\end{table}

\subsection{AL Simulation Experiment}
For both of the AL Simulation settings we experimented in Section~\ref{sec:als}, we leverage the same set of hyper-parameters for fine-tuning our dual-model AL system: $batch\_size\_per\_GPU=2; learning\_rate=1e^{-4}; M_{EG}\_train\_epoch = 20, M_{P}\_train\_epoch = 250; input\_max\_length=512; target\_max\_length=64$ 

\section{Proposal for an Interactive System}
\label{app:system_arch_future}

Our proposed dual-model system can be easily implemented as an interactive human-centered AI system for supporting domain experts and human annotators in labeling both labels and explanations.

\begin{figure}[!h]
    \centering
    \includegraphics[clip,width=.98\columnwidth]{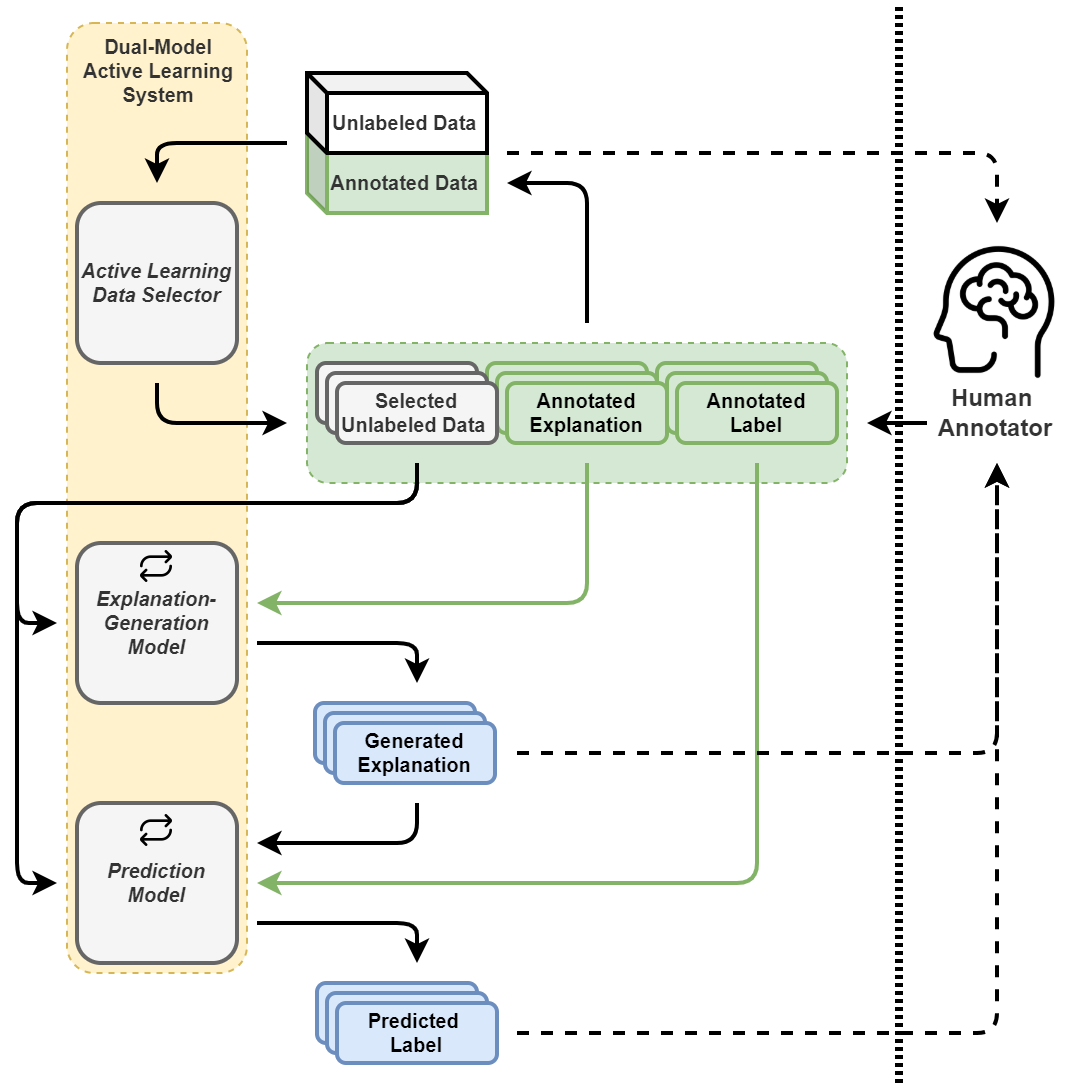}%
    
    \caption{ Our proposed dual-model system can be implemented as an interactive AL-based data annotation system to speed up users' annotation productivity. Such a system can simply have an interface with four output functions (i.e., display unlabeled data, display AL selected data, display generated-explanation, and display predicted labeled) and one input function (i.e., annotate label and explanation for the unlabeled data.}

    \vspace{-1.5em}
    \label{fig:system_arch_future}
\end{figure}

\end{document}